%% file: main.tex
\documentclass{article}
\usepackage[final,nonatbib]{neurips_2025}
\usepackage[utf8]{inputenc}
\usepackage[square,sort,comma,numbers]{natbib}

\title{\bf Pass@K Policy Optimization: \\ Solving Harder Reinforcement Learning Problems}

\author{%
  Christian Walder \& Deep Karkhanis \\
  Google DeepMind \\
  \texttt{cwalder}/\texttt{dkarkhanis}\texttt{@google.com}
}

\date{\today}

\usepackage{booktabs}
\usepackage{listings}
\usepackage{hyperref}
\usepackage{natbib}
\usepackage{graphicx}
\usepackage{amsmath}
\usepackage{amssymb}
\usepackage{amsfonts}
\usepackage{amsthm}
\usepackage{bm}
\usepackage{mathtools}
\usepackage{xcolor}
\usepackage[noabbrev,capitalize]{cleveref}
\usepackage{subcaption}

\definecolor{codegreen}{rgb}{0,0.6,0}
\definecolor{codegray}{rgb}{0.5,0.5,0.5}
\definecolor{codepurple}{rgb}{0.58,0,0.82}
\definecolor{backcolour}{rgb}{0.95,0.95,0.92}

\lstdefinestyle{mystyle}{
    backgroundcolor=\color{backcolour},
    commentstyle=\color{codegreen},
    keywordstyle=\color{magenta},
    numberstyle=\tiny\color{codegray},
    stringstyle=\color{codepurple},
    basicstyle=\ttfamily\footnotesize,
    breakatwhitespace=false,
    breaklines=true,
    captionpos=b,
    keepspaces=true,
    numbers=left,
    numbersep=5pt,
    showspaces=false,
    showstringspaces=false,
    showtabs=false,
    tabsize=2
}

\newtheorem{theorem}{Theorem}
\newtheorem{corollary}{Corollary}
\newtheorem{lemma}{Lemma}
\newcommand{\maxgk}{\ensuremath{\mathrm{max}_{\hspace{-0.05em} g \hspace{-0.1em}}\mathrm{@k}}}
\newcommand{\maxgkmo}{\ensuremath{\mathrm{max}_{\hspace{-0.05em} g \hspace{-0.1em}}\mathrm{@(k-1)}}}
\newcommand{\passone}{\ensuremath{\mathrm{pass@1}}}
\newcommand{\passk}{\ensuremath{\mathrm{pass@k}}}
\newcommand{\passn}{\ensuremath{\mathrm{pass@n}}}
\newcommand{\passkeval}{\ensuremath{\mathrm{pass@k^{eval}}}}
\newcommand{\passkopt}{\ensuremath{\mathrm{pass@k^{opt}}}}
\newcommand{\keval}{\ensuremath{k^\mathrm{eval}}}
\newcommand{\kopt}{\ensuremath{k^\mathrm{opt}}}
\newcommand{\kannealed}{\ensuremath{k^\mathrm{annealed}}}
\newcommand{\gemma}{\textsc{Gemma2}}
\newcommand{\gemmatwob}{\textsc{Gemma2-2B}}
\newcommand{\gemmanineb}{\textsc{Gemma2-9B}}
\newcommand{\llama}{\textsc{Llama3.1}}
\newcommand{\llamaeightb}{\textsc{Llama3.1-8B}}
\newcommand{\pkpo}{\texttt{PKPO}}
\newcommand\arcagione{\texttt{ARC-AGI-1}}
\newcommand{\entropyreg}{\texttt{EntropyReg}}
\newcommand{\mbpp}{\texttt{MBPP}}
\newcommand{\humaneval}{\textsc{HumanEval}}
\newcommand{\mathdata}{\texttt{MATH}}

\input{additional_experiments}

\begin{document}

\maketitle
\begin{abstract}

\qquad Reinforcement Learning algorithms commonly sample multiple ($n>1$) solution attempts for each problem and reward them independently. This optimizes for \passone\ performance and prioritizes individual sample performance over the diversity and collective utility of a set of samples. Such algorithms under-utilize the sampling capacity, limiting exploration and eventual improvement on harder examples. As a fix, we propose \textit{Pass-at-$k$ Policy Optimization} (\pkpo ), a multivariate transformation on batches of rewards which leads to direct optimization of \passk\ performance, thus optimizing for sets of samples that feature a large maximum reward when considered jointly. Our primary contribution is to derive novel low variance unbiased estimators for the \passk\ and its gradient, in both the binary and continuous reward settings. We show that optimizing with these estimators reduces to reinforcement learning with (batches of) rewards that have been jointly transformed by a function that is stable and efficient to compute.

\qquad While previous efforts propose transformations for $k=n$, our transformations are the first to enable robust optimization of the \passk\ for any arbitrary $k \leq n$. Rather than simply trading off \passone\ performance for \passk\ gains, our method allows annealing $k$ during training, optimizing both metrics and often achieving strong \passone\ performance alongside significant \passk\ gains.

\qquad We validate our transformations on illustrative toy experiments, which reveal the variance reducing properties of our formulations. We also include real-world examples using the open-source models \gemma\ and \llama . We find that our transformation effectively optimizes for the target $k$. Furthermore, higher $k$ values enable solving more and harder problems, while annealing $k$ boosts both the \passone\ and \passk . Crucially, for challenging task sets where conventional \passone\ optimization stalls, our \passk\ approach unblocks learning, likely by improving exploration through the prioritization of joint utility over the utility of individual samples

\end{abstract}
\section{Introduction}
Recent years have seen the rapid rise of large language models (LLMs) trained with internet-scale pretraining data \cite{pretraining} with post training using both supervised fine-tuning \cite{posttrainingsft} and reinforcement learning (RL) \cite{gpt4tr,gemini2023,claude,deepseekr1}. The seminal paradigm of RL with human feedback \cite{rlhf} is limited by the human-derived data it is based on and the reward hacking issues that arise from the use of subjective signals more generally \cite{rewardhacking}. To enable progress toward superhuman capabilities, current work is focusing on grounded reward signals that are free of fine-grained human input as in code generation \cite{shojaee2023execution,le2022coderl,dou2024stepcoder,yu2023b,gehring2024rlef} and mathematics \cite{alphacode,alphaproofblog,alphageometry2,yang2023leandojo}.

The policy gradients family of RL methods \cite{policygradients} has proven effective in language model training \cite{deepseekr1}. To scale to new capabilities, model training needs to tackle challenging RL task sets with no known solutions but for which correctness may be verified, as in formal mathematics environments \cite{alphaproofblog}. In such settings, the RL training loop both updates model parameters and searches for solutions to problems at the continuously advancing frontier of model capabilities.

The specific search algorithm introduces a coupling between inference and model updates, which means that naively optimizing the expected single sample reward, or \passone\ may be sub-optimal. While various inference-time search methods are possible \cite{inftimea,inftimeb,inftimec,inftimed}, simply taking multiple independent samples from the model has proven rather effective \cite{iidbeatsselfrepair}. Our contribution is to couple this simple search method with model parameter updates by enabling robust optimization of the \passk\ objective, which is the expected maximum reward over $k$ independent samples.
\input{figures_tex/passgktoy}
\subparagraph{Related Literature}

 The \passk\ was championed by \cite{chen2021codex} who gave a popular unbiased estimator of the metric which we derive from a new perspective (to set up our gradient estimators) as \cref{thm:unbiasedrho}, generalize to continuous rewards in \cref{thm:unbiasedrhog}, and provide additional characterisations in Corollaries~\ref{thm:ngeqkforunbiasedness}~and~\ref{thm:variance}. The concept of expected-maximum optimization under state resetting was proposed by \cite{koyamada2022remax}\footnote{See also the extensions \cite{nishimori2026emergence, takashiro2026advantage, parmas2026ordergrad, nishimori2026retry, tong2026finite}.} as a mechanism to induce exploration. They provided an unbiased policy gradient for the expected-maximum (\maxgk) objective with a rudimentary baseline for variance reduction. Moreover, they showed that combined with an uncertain reward function, optimizing their objective naturally causes the optimal policy to be an exploratory stochastic policy. Concurrently with our work \cite{tangloo} offered an elegant variance reduction method for the gradient of the \passk\ which corresponds to the special case $n=k$ of our \cref{eqn:sloominusone}. Interpreting \passk\ in terms of a partial sort, \cite{differentiable_ranking_and_sorting_with_ot} and \cite{differentiable_ranking_and_sorting_with_ot_xie} present elegant approximations that are rather general but less efficient in our setting.  Others have provided variational approximations for handling the closely related \textit{Best-of-$N$} \cite{bonaware,variationalbon} and other more general \cite{infalign} inference-time algorithms. The contrasting idea of training a model to approximate the \textit{Best-of-$N$} prediction with a single sample was addressed by \cite{bond}. Our contribution can be interpreted as a generalization and variance reduction of \cite{tangloo}. For a general discussion of gradient estimation, variance reduction, and Monte Carlo, we recommend \cite{shakirreview,mcbook}.

\subparagraph{Overview and Contributions}

Our theme is constructing robust estimators of the \passk\ and its gradient given $n\geq k$ samples by averaging (over all $\binom{n}{k}$ subsets of size $k$) simple estimators that are functions of $k$ samples. This is straightforward for binary rewards (\cref{sec:binary}), using the counting proof of \cref{thm:sunbiased}. We generalize to continuous rewards using the key trick of assuming without loss of generality that the rewards are sorted, as in \cref{sec:continuous}. Finally, we give baselining methods that require more involved derivations due to averaging over all subsets that do not include a given element (to retain unbiasedness) but which boil down to the same easy-to-apply results in \cref{sec:variancereduction}, yielding our \textit{Pass-at-$k$ Policy Optimization} (\pkpo ). We present toy experiments in \cref{sec:toyexperiments} which demonstrate the variance reduction afforded by our estimators. Finally, \cref{sec:gemmaexperiments} demonstrates that using our reward transformation solves more tasks and selectively optimizes \passk\ through RL experiments on \gemma\ \cite{gemma2} and \llama\ \cite{llama}, showcasing real-world impact.

\subparagraph{How to Apply this Method}
It is easy to adapt any policy gradient algorithm to use our results. Assume a vector $(g(x_1), g(x_2), \dots , g(x_n))^\top$ of per-sample rewards for a given task. For example, the $x_i$ could be model samples of source code addressing a specific task (which should be the same for all $n$ samples), and $g$ could provide a numeric score that measures how many tests the code passes, or an overall binary pass indicator, or some combination with additional stylistic or brevity terms, \textit{etc.} Then in order to optimize the \passk\ of \cref{eqn:def:passatk} (or the continuous analog \maxgk\ of \cref{eqn:def:passgk}) we simply transform the vector of rewards using either the \texttt{sloo} or the \texttt{sloo\_minus\_one} function of \cref{code:listing}, which map $\mathbb R^n\mapsto \mathbb R^n$.\footnote{While our experiments focus on \texttt{sloo\_minus\_one}, we recommend experimenting with both estimators.} 
\section{Binary Rewards}
\label{sec:binary}
Given a binary reward function $f:\mathcal X \rightarrow \{0,1\}$ on the action space $\mathcal X$, the \passk\ for the model $p(x|\theta)$ is the probability that at least one of $k$ samples drawn i.i.d. is correct:
\begin{align}
    \label{eqn:def:passatk}
    \passk
    & =
    \mathbb P\left[ \bigvee_{i=1}^k \left[f(x_i)=1\right] \right]
    \\
    & =
    \mathbb E \left[1 - \prod_{i=1}^k\left(1-f(x_i)\right) \right],
\end{align}
where the expectation is over i.i.d. $x_1, x_2, \dotsm, x_k \sim p(x|\theta)$.
\subsection{An Unbiased \passk\ Estimator}
An estimator for the \passk\ was given in \cite{chen2021codex}: given $n \geq k$ i.i.d. samples of which $c$ are correct, the estimator is
\begin{align}
    \label{eqn:rho}
    \rho(n,c,k) \equiv 1 - \frac{\binom{n-c}{k}}{\binom{n}{k}}.
\end{align}
The following was proven in \citep{chen2021codex}; we give a different proof that sets up our gradient estimator.
\begin{theorem}
\label{thm:unbiasedrho}
$\rho(n,c,k)$ is an unbiased estimator of the \passk .
\end{theorem}
\proof{
Let $x_1, x_2, \dotsm, x_n \sim p(x|\theta)$, $f_i=f(x_i)$, and $\mathcal I$ be a set of $k$ elements sampled uniformly without replacement from $\{1,2,\dots,n\}$. Then
\begin{align}
    \passk = \mathbb E_{x_1, x_2, \dots, x_n} \mathbb E_{\mathcal I} \left[ 1-\prod_{i\in\mathcal I} (1-f_i) \right].
\end{align}
Averaging over all subsets of size $k$ recovers $\rho$:
\begin{align}
    \frac{1}{\binom{n}{k}} \sum_{\substack{|\mathcal I| = k \\ \mathcal I \subseteq \{1,2,\dots,n\}}} \left(1-\prod_{i\in\mathcal I} (1-f_i)\right)
    & = \label{eqn:suma}
    1 - \frac{1}{\binom{n}{k}} \sum_{\substack{|\mathcal I| = k \\ \mathcal I \subseteq \{1,2,\dots,n\}}} \prod_{i\in\mathcal I} (1-f_i) \\
    & = \label{eqn:binoma}
    1 - \frac{\binom{n-c}{k}}{\binom{n}{k}}
    \\
    & \equiv
    \rho(n, c, k),
\end{align}
where \eqref{eqn:binoma} holds because the sum on the r.h.s. of \eqref{eqn:suma} is the number of subsets of size $k$ of the $(n-c)$ incorrect elements. Since averaging in this way retains unbiasedness, this completes the proof. \qed
}
We show in \cref{thm:ngeqkforunbiasedness} that no such unbiased estimator exists for $n < k$, and in \cref{thm:variance} that the asymptotic variance of this estimator decreases at a rate of $1/n$.
\subsection{An Unbiased \passk\ Gradient Estimator}
\input{figures_tex/passgkm}
Given a mini-batch of $n$ i.i.d. samples $x_1, x_2, \dots, x_n$ from $p(x|\theta)$ with corresponding correctness labels $f_i \in \{0,1\}$, we want to optimize the \passk\ w.r.t. the model parameters $\theta$. Letting $c=\sum_{i=1}^n f_i$ be the number of correct samples, we will demonstrate unbiasedness of the estimator
\begin{align}
\label{eqn:g:estimator}
    \widehat{\nabla}
    =
    \sum_{i=1}^n r_i \nabla_\theta \log p(x_i|\theta), 
    ~~~\textrm{where}~~~
    r_i =
    \begin{cases}
    \frac{k}{n} & \mathrm{if~} f_i=1 \\
    \frac{k}{n} \cdot \rho(n-1, c, k-1) & \mathrm{if~} f_i=0, \\
    \end{cases}
\end{align}
that assigns more weight to correct samples, while also assigning some reward to incorrect samples to encourage exploration.
The following well-known results will be used to show that \eqref{eqn:g:estimator} is unbiased.
\begin{lemma}[Policy Gradients]
\label{thm:pg}
For any absolutely continuous distribution $p(x|\theta)$
\begin{align}
    \mathbb E_{x\sim p(x|\theta)} \left[ r(x) \nabla_\theta \log p(x|\theta)\right]
    = 
    \nabla_\theta \mathbb E_{x\sim p(x|\theta)} \left[ r(x) \right].
\end{align}
\end{lemma}
\begin{corollary}
\label{thm:constantpg}
If $c$ is constant w.r.t. both $\theta$ and $x$ then
$\mathbb E_{p(x|\theta)} \left[ c \nabla_\theta \log p(x|\theta) \right] = 0$.
\end{corollary}
\proof{By \cref{thm:pg}, $\mathbb E_{p(x|\theta)} \left[ c \nabla_\theta \log p(x|\theta) \right]=\nabla_\theta \mathbb E\left[c\right] = \nabla_\theta c = 0$. \qed
}

We can now give our first main result:
\begin{theorem}
\label{thm:unbiasedrhograd}
$\widehat{\nabla}$ is an unbiased estimator of the gradient of the \passk :
\begin{align}
    \mathbb E_{x_1, x_2, \dots, x_n\sim p(x|\theta)} \left[\widehat \nabla \right] = \nabla_\theta \passk .
\end{align}
\end{theorem}
See \cref{sec:binaryproof} for a proof.
\section{Continuous Rewards}
\label{sec:continuous}
We generalize the \passk\ to non-binary rewards $g:\mathcal X \rightarrow \mathbb R$ as
\begin{align}
    \label{eqn:def:passgk}
    \maxgk \equiv \mathbb E \left[\max \left( \left\{ g(x_i) \right\}_{i=1}^k\right) \right].
\end{align}
\subsection{An Unbiased \texorpdfstring{\maxgk}{maxg@k} Estimator}
The following estimator for the \maxgk\ is a direct analog of $\rho$: given $n \geq k$ i.i.d. samples, assuming w.l.o.g. that the rewards $g_i=g(x_i)$ are sorted, so that $g_1 \leq g_2\leq\dots\leq g_n$ the estimator is
\begin{align}
\label{eqn:g:estsum}
    \rho^{(g)}(n,c,k) & \equiv \frac{1}{\binom{n}{k}} \sum_{i=k}^n \mu_i g_i, \\
\shortintertext{where}
\mu_i & = \binom{i-1}{k-1}.
\end{align}
To compute this stably we cancel factors in the binomial coefficients to get\footnote{\label{footnote:ruixu}We thank to Ruixu Zhou of Tsinghua University for correcting errors in equations \ref{eqn:stablerhog}, \ref{eqn:bsum} and \ref{eqn:bsum2}.}
\begin{align}
\label{eqn:stablerhog}
    \rho^{(g)}(n,c,k) \equiv \frac{k}{n-k+1} \sum_{i=k}^n g_i
    \prod_{j=1}^{k-1} \frac{i-j}{n-j+1}.
\end{align}
\begin{theorem}
\label{thm:unbiasedrhog}
$\rho^{(g)}(n,c,k)$ is an unbiased estimator of the \maxgk .
\end{theorem}
\proof{The proof is similar to \cref{thm:unbiasedrho}. Here we exploit the assumption that the $g_i$ are sorted, so
\begin{align}
    \frac{1}{\binom{n}{k}} \sum_{\substack{|\mathcal I| = k \\ \mathcal I \subseteq \{1,2,\dots,n\}}} \max_{i\in \mathcal I} g_i
    & = 
    \frac{1}{\binom{n}{k}} \sum_{\substack{|\mathcal I| = k \\ \mathcal I \subseteq \{1,2,\dots,n\}}} g_{\max_{i\in \mathcal I}}
    \\
    \label{eqn:g:mu}
    & =
    \frac{1}{\binom{n}{k}} \sum_{i=k}^n \mu_i g_i
    \\
    & \equiv
    \rho^{(g)}(n, c, k),
\end{align}
since $\mu_i$ is the number of subsets of $1,2,\dots,i-1$ of size $k-1$, which equals $\binom{i-1}{k-1}$. The sum starts at $k$ because all subsets of size $k$ include elements that are greater than or equal to $g_k$.
\qed
}

See Line~\ref{code:rho} of \cref{code:listing} for an implementation of $\rho^{(g)}$.
\subsection{An Unbiased \texorpdfstring{\maxgk}{maxg@k} Gradient Estimator}
\label{sec:unbaselinedpassgk}
We propose the gradient estimator
\begin{align}
\label{eqn:estimator}
    \widehat{\nabla}^{(g)}
    & =
    \sum_{i=1}^n {s_i} \nabla_\theta \log p(x_i|\theta),
\end{align}
where if we assume w.l.o.g. that the $g_i$ are sorted, the $s_i$ are a weighted combination of them,
\begin{align}
    \label{eqn:ssum}
    s_i & = \frac{1}{\binom{n}{k}} \sum_{j=i}^n m_{ij} g_j,
\shortintertext{where the diagonals are}
    \label{eqn:g:mii}
    m_{ii} & = 
    \begin{cases}
    \binom{i-1}{k-1} & \mathrm{~if~} i \geq k-1 \\
    0 & \mathrm{otherwise,}
    \end{cases}
\shortintertext{and the off-diagonals are}
    \label{eqn:g:mij}
    m_{ij} & = 
    \begin{cases}
    \binom{j-2}{k-2}
    &
    \mathrm{~if~} (j > i) \wedge (j \geq k) \wedge (k\geq 2) \\
    0
    &
    \mathrm{otherwise.}
    \end{cases}
\end{align}
\begin{theorem}
$\widehat{\nabla}^{(g)}$ is an unbiased estimator of the gradient of the \maxgk :
\label{thm:sunbiased}
\begin{align}
    \mathbb E_{x_1, x_2, \dots, x_n\sim p(x|\theta)} \left[\widehat{\nabla}^{(g)} \right] = \nabla_\theta \maxgk .
\end{align}
\end{theorem}
\proof{
The proof is analogous to that of  \cref{thm:unbiasedrhograd}. Here we have
\begin{align}
    \widehat \nabla 
    & \equiv
    \rho^{(g)}(n,c,k) \nabla_\theta \sum_{i=1}^n \log p(x_i|\theta)
    \\
    \label{eqn:g:substitutesuma}
    & =
    \frac{1}{\binom{n}{k}} \sum_{\substack{|\mathcal I| = k \\ \mathcal I \subseteq \{1,2,\dots,n\}}} \max_{j\in\mathcal I} g_j
    \sum_{i=1}^n \nabla_\theta \log p(x_i|\theta)
    \\
    \label{eqn:g:implicitm}
    & \stackrel{\mathbb E}{\equiv}
    \frac{1}{\binom{n}{k}} \sum_{i=1}^n \nabla_\theta \log p(x_i|\theta) \sum_{j=1}^n m_{ij} g_j,
\end{align}
By assumption the $g_i$ are sorted, so $\max_{j\in\mathcal I} g_j =  g_{\max_{j\in\mathcal I}}$. Therefore $m_{ij}$ is the number of subsets $\mathcal I$ of $\{1,2,\dots,n\}$ that
\begin{enumerate}
    \item are of size $k$,
    \item have $j\geq i$ as the largest element (so that we can factor out $g_j$),
    \item include $i$ (so that \eqref{eqn:g:implicitm} holds in expectation by \cref{thm:constantpg}).
\end{enumerate}
Due to the second condition, the form of $m_{ij}$ depends on whether $i=j$.

The diagonals $m_{ii}$ are zero if $i<k$ since the largest element of any subset of size $k$ is at least $k$. If $i\geq k$ then we fix $i$ and are left with $i-1$ elements from which to choose $k-1$ which we can do  $\binom{i-1}{k-1}$ ways in line with \eqref{eqn:g:mii}.

The $m_{ij}$ for $i\neq j$ are obtained by fixing $i$ and $j$ leaving $j-2$ elements $1, 2, \dots, i-1, \dots, i+1, \dots, j-1$ from which to choose $k-2$ which we can do $\binom{j-2}{k-2}$ ways in line with \eqref{eqn:g:mij}.
\qed
}
\begin{theorem}
\label{thm:stime}
$s_1, s_2, \dots, s_n$ can be computed in total time $\mathcal O(k+n\log n)$.
\end{theorem}
See \cref{sec:passgktimeproof} for the proof and Line~\ref{code:s} of \cref{code:listing} for an implementation based on it.
\section{Variance Reduction}
\label{sec:variancereduction}
\subsection{Leave-One-Out Baseline for the Simple Case}
A popular variance reduction method \cite{shakirreview,mcbook,deepseekr1} for point-wise rewards $g(x)$ subtracts the mean of the leave one out (LOO) rewards within each mini-batch $x_1, x_2, \dots, x_n$:
\begin{align}
    \label{eqn:basicloo}
    g^{\mathrm{(loo)}}(x_i) = g(x_i) - \frac{1}{n-1}\sum_{\substack{j=1\\ j\neq i}}^n g(x_j).
\end{align}
Since the subtracted part does not depend on $x_i$, by \cref{thm:constantpg} this retains unbiasedness.
\subsection{Leave-One-Out Baseline for \texorpdfstring{\maxgk}{}}
\label{sec:loomaxgk}
Baselining the $s_i$ of \eqref{eqn:ssum} in this way introduces bias, however, as each $s_i$ depends on all $x_1, \dots , x_n$.
We instead apply LOO to the following form of $s_i$ that follows from \cref{thm:sunbiased} and the proof thereof:
\begin{align}
    \label{eqn:salternate}
    s_i 
    & =
    \frac{1}{\binom{n}{k}} \sum_{\substack{|\mathcal I| = k \\ i \in \mathcal I \\ \mathcal I \subseteq \{1,2,\dots,n\}}} \max_{j\in\mathcal I} g_j \\
    \label{eqn:bigsdef}
    & \equiv S(i, k, \{1,2,\dots, n\}),
\end{align}
We can then retain unbiasedness by excluding $i$ from the baseline, by defining
\begin{align}
\label{eqn:sloo}
    s_i^{\mathrm{(loo)}}
    & 
    \equiv 
    S(i, k, \{1,2,\dots, n\}) -
    \frac{1}{n-1}
    \sum_{\substack{j=1\\ j\neq i}}^n S(j, k, \{1,2,\dots, n\} \setminus i)
    .
\end{align}
\begin{theorem}
$s_1^{(loo)}, s_2^{(loo)}, \dots, s_n^{(loo)}$ can be computed in total time $\mathcal O(k+n\log n)$.
\end{theorem}
\proof{
Given \eqref{thm:stime} it is sufficient to consider computing, for $i=1,2,\dots,n$,
\begin{align}
    \label{eqn:bdef}
    b_i^{(k)} \equiv \sum_{\substack{j=1\\ j\neq i}}^n S(j, k, \{1,2,\dots, n\} \setminus i).
\end{align}
By assuming w.l.o.g. an ascending ordering of $g(x_i)$, excluding the first index does not change the ordering of the remaining indices. The first term is therefore$^{\ref{footnote:ruixu}}$
\begin{align}
    b_1^{(k)}
     =
    \sum_{i=2}^N s_i
     =
    \label{eqn:bsum}
    \frac{1}{\binom{n-1}{k}}
    \sum_{i=2}^N \big( m_{ii} + m_{i-1,i} (i-2) \big) g(x_i),
\end{align}
where \eqref{eqn:bsum} follows from \eqref{eqn:srecursion}.
From \eqref{eqn:srecursion} we obtain for $1 \leq i < n$ the left to right recursion
\begin{align}
    \label{eqn:bsum2}
    b_{i+1}^{(k)}
    & =
    b_i^{(k)} +
    \frac{1}{\binom{n-1}{k}}
    \big(
    g(x_i)-g(x_{i+1})
    \big)
    \big(
    m_{ii} + m_{i-1,i} (i-2)
    \big).
\end{align}
Similar arguments to the proof of \cref{thm:stime} therefore imply the same time complexity.
\qed
}

Line~\ref{code:sloo} of \cref{code:listing} implements $s_i^{(\mathrm{loo})}$ using the recursion in the above proof.

\subsection{\texorpdfstring{\maxgkmo}{} Leave-One-Out Baseline for \texorpdfstring{\maxgk}{}}
\label{sec:loomaxgkminusone}
The baseline $b_i^{(k)}$ is an average of the $\maxgk$ estimates over sets of size $k$. For a number of samples $n$ equal to $k$, there are no such subsets to construct the baseline. \cite{tangloo} recently overcame this issue for the specific case $n=k$ by using $\maxgkmo$ as the baseline statistic. We generalize their approach to $k<n$ and to averaging over all subsets by defining similarly to \cref{eqn:salternate}
\begin{align}
    \label{eqn:sloominusone}
    s_i^{(\mathrm{loo-1})}
    & =
    \frac{1}{\binom{n}{k}} \sum_{\substack{|\mathcal I| = k \\ i \in \mathcal I \\ \mathcal I \subseteq \{1,2,\dots,n\}}} \big( \max_{j\in\mathcal I} g_j - \max_{b\in\mathcal I \setminus i} g_b\big).
\end{align}
Averaging smaller but more numerous subsets in the baseline reduces variance but introduces bias (in the baseline, not $s_i^{(\mathrm{loo}-1)}$).
Given our previous results it is straightforward to show
\begin{theorem}
$s_1^{(loo-1)}, s_2^{(loo-1)}, \dots, s_n^{(loo-1)}$ can be computed in total time $\mathcal O(k+n\log n)$.
\end{theorem}
\proof{
By the linearity of the expectation we can split the two terms in the parentheses of \cref{eqn:sloominusone} into two separate sums. The first summation is by definition simply $s_i$ of \cref{eqn:ssum}. The (negation of the) second summation can be computed efficiently using
\begin{align}
    \label{eqn:bminusoneneatform}
    \frac{1}{\binom{n}{k}} \sum_{\substack{|\mathcal J| = k \\ \mathcal J \subseteq \{1,2,\dots,n\}}} \max_{b\in\mathcal J \setminus i} g_b
     = 
    \frac{1}{\binom{n}{k}} \sum_{\substack{|\mathcal B| = k-1 \\ \mathcal B \subseteq \{1,2,\dots,n\}\setminus i}} \max_{b\in\mathcal B} g_b
     = \frac{k}{n(k-1)} b_i^{(k-1)},
\end{align}
where the final equality follows with a little algebra from \cref{eqn:bigsdef} and \cref{eqn:bdef}. \qed
}
\cref{code:listing} implements $s_i^{(\mathrm{loo-1})}$ using \eqref{eqn:bminusoneneatform}; \cref{fig:baseline} compares $s_i$, $s_i^{(\mathrm{loo})}$, and $s_i^{(\mathrm{loo}-1)}$.
\section{Experiments}
\label{sec:experiments}
\subsection{One-Dimensional Toy Example}
\label{sec:toyexperiments}
We start with a policy that is Gaussian with a fixed standard deviation and mean parameter $\theta$ we wish to learn, so that $x\sim \mathcal N(\theta, 0.1)$. We set the raw reward to be
\begin{align}
    g(x) =
    \begin{cases}
    x^2 & 0 \leq x \leq 1 \\
    0 & \mathrm{otherwise}.
    \end{cases}
\end{align}
The optimal policy under the \maxgk\ reward varies with $k$ (see \cref{fig:passgktoy}). The variance of our estimators is compared in \cref{fig:toyvariance} where $s^{(\textrm{loo}-1)}$ is the strongest.
\gemmaninebmath
\llamaeightbmath
\gemmaninebcoding
\llamaeightbcoding
\subsection{RL on Open Source LLMs}
\label{sec:gemmaexperiments}
We demonstrate promising RL results with the \texttt{2B} and \texttt{9B} parameter variants of \gemma\ \cite{gemma2} and the \texttt{8B} parameter variant of \llama\ on real-world problems in MATH \cite{math_12000}, code generation \cite{austin2021programsynthesislargelanguage} \cite{chen2021evaluatinglargelanguagemodels}, and the easy public subset of \arcagione\ \cite{arc_agi_1}. The latter is a challenging reasoning task-set even for state-of-the-art models much larger than \gemma .

For \gemmatwob\ we use a \texttt{v5litepod-128} \cite{gcp} which needs around 4 hours per 1000 training steps. Each RL training run \cite{ppo_rl} involves sampling a fixed $n$ number of completions $\{x_i\}_{i=1}^n$ for a given prompt at a given training step. For our experiments, we set $n=16$. The rewards are computed for every completion using a reward function $g(\cdot)$. We transform these rewards $\{g(x_i)\}_{i=1}^n$ using our unbiased estimator $s^{(\mathrm{loo}-1)}$ of \eqref{eqn:sloominusone}, which we favour due to \cref{fig:toyvariance}, and which we refer to as \pkpo . We repeat the training for a selection of $k^\mathrm{opt}$, thus optimizing a different \passkopt\ each time. Since $k^\mathrm{opt}=1$ leads to no reward transformation, this is our baseline (although we use basic LOO mean centering of \cref{eqn:basicloo}, without which the training diverges). For each run, we measure \passkeval\ for every \keval $\in \{1,2,4,8,12,16\}$ at each step. Additionally, we also track model entropy and cumulative solve rate during training. The latter is defined as the fraction of tasks from the task-set for which the model has sampled a correct solution at least once; this is a critical metric that reflects the success of the model's exploration and measures its ability to find novel solutions.
\gemmaninebarc
\llamaeightbarc

\textbf{Entropy regularization baseline} In addition to our \pkpo\ and the special case thereof of \cite{tangloo}, we also add the entropy regularization baseline, which is PPO with an additional entropy term in the objective. We give this baseline an arguably unfair advantage by performing a small sweep over the values $0.001, 0.005, 0.01, 0.05, 0.1$ for the \texttt{entropy\_coefficient} for each (model, benchmark) pair and only report the best result as \entropyreg .

\subsubsection{Choosing \texorpdfstring{\kopt\ }{k opt} selectively optimizes \texorpdfstring{\passkeval\ }{k eval} and solves more tasks}

We use the training split of Hendrycks MATH \cite{math_12000} which contains 12,000 problems as our task set. \cref{fig:cumu_solves} shows that a higher \kopt\ in our transformation leads to a consistently higher cumulative solve rate throughout training, as well as a higher entropy. By optimizing \passk\ instead of \passone , the model appears to better utilize the exploration budget thus finding more solutions.

In \cref{fig:rolling_passk}, we compare \passkeval\ across our runs (\kopt\ $\in \{1,4,8\}$) for various \keval . We find the best \passkeval\ when $k^{\mathrm{opt}}=k^\mathrm{eval}$ (or $k^{\mathrm{opt}}$ is closest to $k^\mathrm{eval}$  among available $k^\mathrm{opt}$). Non-transformed rewards optimize \passone , leading to sub-optimal \passkeval\ for \keval $\neq 1$, and the deficit worsens as $k^\mathrm{eval}$ increases.
Thus, our experiments also demonstrate that setting \kopt $\coloneqq$ \keval\ in our transformation suffices to optimize \passkeval\ for a \keval $\leq n$. This generalizes the already powerful result of \cite{tangloo} by alleviating the coupling that restricts to optimizing either \passn\ or \passone .
In other words, since RL training of LLMs typically samples a large batch ($n \gg 1$), failing to use our transformation results in sub-optimal \passk\ performance, especially for modest values of $k$.

As $k^{\mathrm{opt}} \longrightarrow n$, the variance of our estimator increases as there are fewer subsets in \eqref{eqn:sloominusone} (see \cref{fig:toyvariance}). We presume this is why 1) gains of $k^\mathrm{opt}=8$ over $k^\mathrm{opt}=4$ are more prominent when \keval\ $\in \{12,16\}$ than when \keval\ $=8$. That is, when $k^\mathrm{eval}$ is further away from $k^\mathrm{opt} \in \{4,8\}$ than when it is closer, and 2) the special case $n=k^{\mathrm{opt}}$ of \cite{tangloo} struggles to optimize the \passn .

\subsubsection{\pkpo\ robustly improves \texorpdfstring{\passk}{pass at k} on held out evaluations}
Tables~\ref{tab:gemma-9b-final}-\ref{tab:llama-31-8b-coding} above (and Tables~\ref{tab:gemma-2b-final}-\ref{tab:gemma-2b-coding} in the appendix) present performance on held-out sets for two tasks. We report the mean and standard error based on three runs with different random seeds. For math, we train on the train split and evaluate on the test split of Hendrycks MATH \cite{math_12000}. \nocite{mathreason}
To evaluate coding, we use \mbpp\ \cite{austin2021programsynthesislargelanguage} for training and evaluate on \humaneval\ \cite{chen2021evaluatinglargelanguagemodels}. \mbpp\ has multiple unit tests per problem and hence we use this not only as a proxy for additional benchmarks but also to showcase our handling of a continuous reward function (\% unit tests passed).

\subsubsection{Improving \texorpdfstring{\passk}{pass at k eval} without sacrificing \texorpdfstring{$\mathrm{pass}@1$}{pass at 1}}
\cref{fig:annealk} demonstrates that as \pkpo\ can use any arbitrary $\kopt \leq n$, this allows varying $k^\mathrm{opt}$ over the course of training to good effect. We show a simple annealing procedure which starts training with a high $k^\mathrm{opt} = 8$ and reduces it to $k^\mathrm{opt} = 1$ after 1500 steps. This trains the model to initially prioritize exploration (optimize \passk) and then consolidate the single-sample policy (optimize \passone). This switch is apparent in \cref{fig:annealk_passone}, at step 1500 where the slope of \kannealed\ changes. While traditional methods like \cite{tangloo} suffer from a trade-off between \passk\ and \passone , we get a final model which has higher \passkeval\ for all \keval$ > 1$ with no sacrifice in \passone . 

\subsubsection{\pkpo\ is essential for learning on hard problems}
\cref{fig:arc_agi} shows the limitation of traditional \passone\ optimization through RL on an especially challenging task-set. We use the easy subset of \arcagione\ \cite{arc_agi_1}. We observe that conventional \passone\ optimization stalls. However, our \passk\ approach unblocks learning, and results in higher \passkeval\ across all \keval\ including \keval $=1$. Furthermore, we see higher \kopt\ leads to more effective and faster learning. This is likely because the benefits of prioritizing joint utility over individual sample utility are more prominent on a harder task-set.
\input{figures_tex/anneal}

Tables~\ref{tab:gemma-9b-arc}~and~\ref{tab:llama-31-8b-arc} show more extensive experiments on \arcagione. We make an 80:20 train:test split of the same easy subset as before and report the cumulative solve rate on the train set and \passk\ rate on the test set. We train to saturation (no change in cumulative rate for 1k steps), and again use three random restarts to provide standard errors. By encouraging exploration in a direct and stable manner, our method unblocks learning unlike other methods. Entropy Regularization does indeed sacrifice \texttt{pass@1} and slightly improves \passk\ by promoting exploration, but it is hard to tune, and is significantly outperformed by our method. Moreover, it has no explicit way to optimize for a specific \texttt{k\_eval}. \cite{tangloo} targets the same objective as \pkpo , but couples the minibatch size to $k$ and thereby incurs higher variance than \pkpo\ with $k<n$.
\section{Conclusions and Outlook}
In RL training with multiple independent samples per task, optimizing the \passk\ maximizes the expectation of the \textit{best} reward in the set of samples, rather than the \textit{average} one. This preserves model output diversity, which leads to solving more problems and ultimately yields stronger policies. We provide drop-in replacements for more traditional RL reward transformations that robustly and efficiently optimize the \passk . This work can be extended in various ways, such as to other inference-time search algorithms, and to more sophisticated baseline techniques.
\clearpage
\bibliographystyle{alpha}
\bibliography{references}
\clearpage
\appendix
\section{Additional Theoretical Statements and Proofs}
\input{ngeqkthm}
\input{variance}
\input{binaryproof}
\input{passgktimeproof}
\clearpage
\section{Implementation}
\input{figures_tex/python}
\clearpage
\section{Additional Figures}
\input{figures_tex/toyvariance}
\input{figures_tex/baseline}
\input{figures_tex/cumulative_solve_rate}
\input{figures_tex/rolling_pass_k}
\input{figures_tex/arc_agi}
\gemmatwobmath
\gemmatwobcoding
\clearpage
\input{neurips_checklist}
\end{document}

%% file: additional_experiments.tex
\newcommand{\gemmatwobmath}{
\begin{table}[ht!]
\centering
\caption{Results for \gemmatwob\ on the \mathdata\ benchmark.}
\label{tab:gemma-2b-final}
\begin{tabular}{lccccc}
\toprule
\gemmatwob & \texttt{k\_eval=1} & \texttt{k\_eval=2} & \texttt{k\_eval=4} & \texttt{k\_eval=8} & \texttt{k\_eval=16} \\
\toprule
\texttt{k\_opt=1} & \textbf{15.91} $\pm$ 0.40 & 18.12 $\pm$ 0.43 & 23.37 $\pm$ 0.48 & 29.58 $\pm$ 0.53 & 35.02 $\pm$ 0.58 \\
\texttt{k\_opt=2} & 15.15 $\pm$ 0.41 & \textbf{20.73} $\pm$ 0.45 & 25.81 $\pm$ 0.50 & 31.96 $\pm$ 0.55 & 37.75 $\pm$ 0.60 \\
\texttt{k\_opt=4} & 14.86 $\pm$ 0.43 & 19.86 $\pm$ 0.47 & \textbf{27.59} $\pm$ 0.52 & 34.27 $\pm$ 0.58 & 38.91 $\pm$ 0.62 \\
\texttt{k\_opt=8} & 14.19 $\pm$ 0.46 & 19.33 $\pm$ 0.50 & 26.45 $\pm$ 0.55 & \textbf{35.49} $\pm$ 0.60 & \textbf{40.73} $\pm$ 0.65 \\
\midrule
\cite{tangloo} & 13.11 $\pm$ 0.50 & 18.09 $\pm$ 0.54 & 24.58 $\pm$ 0.60 & 31.81 $\pm$ 0.66 & 37.24 $\pm$ 0.71 \\
\entropyreg & 14.51 $\pm$ 0.48 & 18.95 $\pm$ 0.52 & 25.33 $\pm$ 0.58 & 30.95 $\pm$ 0.64 & 36.18 $\pm$ 0.69 \\
\bottomrule
\end{tabular}
\end{table}
}

\newcommand{\gemmaninebmath}{
\begin{table}[ht!]
\centering
\caption{Results for \gemmanineb\ on the \mathdata\ benchmark.}
\label{tab:gemma-9b-final}
\begin{tabular}{lccccc}
\toprule
\gemmanineb & \texttt{k\_eval=1} & \texttt{k\_eval=2} & \texttt{k\_eval=4} & \texttt{k\_eval=8} & \texttt{k\_eval=16} \\
\toprule
\texttt{k\_opt=1} & \textbf{22.24} $\pm$ 0.50 & 25.35 $\pm$ 0.55 & 30.73 $\pm$ 0.59 & 37.08 $\pm$ 0.64 & 42.59 $\pm$ 0.68 \\
\texttt{k\_opt=2} & 21.46 $\pm$ 0.51 & \textbf{28.61} $\pm$ 0.56 & 32.92 $\pm$ 0.61 & 39.59 $\pm$ 0.66 & 45.34 $\pm$ 0.70 \\
\texttt{k\_opt=4} & 21.25 $\pm$ 0.53 & 27.15 $\pm$ 0.58 & \textbf{34.93} $\pm$ 0.63 & 41.71 $\pm$ 0.69 & 47.05 $\pm$ 0.74 \\
\texttt{k\_opt=8} & 20.69 $\pm$ 0.56 & 26.78 $\pm$ 0.60 & 33.68 $\pm$ 0.66 & \textbf{42.62} $\pm$ 0.72 & \textbf{48.37} $\pm$ 0.77 \\
\midrule
\cite{tangloo} & 19.48 $\pm$ 0.61 & 25.41 $\pm$ 0.67 & 31.17 $\pm$ 0.73 & 39.34 $\pm$ 0.79 & 44.82 $\pm$ 0.83 \\
\entropyreg & 20.85 $\pm$ 0.58 & 26.05 $\pm$ 0.64 & 32.48 $\pm$ 0.70 & 38.21 $\pm$ 0.76 & 43.95 $\pm$ 0.81 \\
\bottomrule
\end{tabular}
\end{table}
}

\newcommand{\llamaeightbmath}{
\begin{table}[ht!]
\centering
\caption{Results for \llamaeightb\ on the \mathdata\ benchmark.}
\label{tab:llama-31-8b-final}
\begin{tabular}{lccccc}
\toprule
\llamaeightb & \texttt{k\_eval=1} & \texttt{k\_eval=2} & \texttt{k\_eval=4} & \texttt{k\_eval=8} & \texttt{k\_eval=16} \\
\toprule
\texttt{k\_opt=1} & \textbf{51.15} $\pm$ 0.61 & 51.82 $\pm$ 0.64 & 53.69 $\pm$ 0.68 & 55.41 $\pm$ 0.72 & 56.83 $\pm$ 0.76 \\
\texttt{k\_opt=2} & 49.72 $\pm$ 0.62 & \textbf{53.51} $\pm$ 0.66 & 55.45 $\pm$ 0.70 & 57.23 $\pm$ 0.74 & 58.71 $\pm$ 0.78 \\
\texttt{k\_opt=4} & 49.18 $\pm$ 0.64 & 52.20 $\pm$ 0.68 & \textbf{57.83} $\pm$ 0.72 & 58.47 $\pm$ 0.77 & 59.28 $\pm$ 0.81 \\
\texttt{k\_opt=8} & 48.63 $\pm$ 0.67 & 52.14 $\pm$ 0.71 & 56.28 $\pm$ 0.75 & \textbf{59.04} $\pm$ 0.80 & \textbf{61.88} $\pm$ 0.84 \\
\midrule
\cite{tangloo} & 48.21 $\pm$ 0.70 & 50.93 $\pm$ 0.75 & 54.38 $\pm$ 0.80 & 57.11 $\pm$ 0.85 & 58.55 $\pm$ 0.90 \\
\entropyreg & 48.51 $\pm$ 0.68 & 51.95 $\pm$ 0.73 & 55.33 $\pm$ 0.78 & 56.95 $\pm$ 0.83 & 58.18 $\pm$ 0.88 \\
\bottomrule
\end{tabular}
\end{table}
}

\newcommand{\gemmatwobcoding}{
\begin{table}[ht!]
\centering
\caption{Results for \gemmatwob\ on the Coding benchmark.}
\label{tab:gemma-2b-coding}
\begin{tabular}{lccccc}
\toprule
\gemmatwob & \texttt{k\_eval=1} & \texttt{k\_eval=2} & \texttt{k\_eval=4} & \texttt{k\_eval=8} & \texttt{k\_eval=16} \\
\toprule
\texttt{k\_opt=1} & \textbf{19.82} $\pm$ 0.53 & 23.81 $\pm$ 0.57 & 29.75 $\pm$ 0.62 & 36.33 $\pm$ 0.66 & 42.04 $\pm$ 0.71 \\
\texttt{k\_opt=2} & 18.70 $\pm$ 0.54 & \textbf{26.94} $\pm$ 0.59 & 33.82 $\pm$ 0.64 & 40.95 $\pm$ 0.69 & 47.03 $\pm$ 0.74 \\
\texttt{k\_opt=4} & 18.69 $\pm$ 0.56 & 26.43 $\pm$ 0.61 & \textbf{36.81} $\pm$ 0.67 & 44.81 $\pm$ 0.73 & 50.55 $\pm$ 0.78 \\
\texttt{k\_opt=8} & 17.94 $\pm$ 0.59 & 25.86 $\pm$ 0.64 & 35.88 $\pm$ 0.70 & \textbf{46.45} $\pm$ 0.77 & \textbf{52.83} $\pm$ 0.83 \\
\midrule
\cite{tangloo} & 16.81 $\pm$ 0.65 & 24.27 $\pm$ 0.69 & 33.11 $\pm$ 0.76 & 41.98 $\pm$ 0.84 & 47.26 $\pm$ 0.89 \\
\entropyreg & 18.05 $\pm$ 0.62 & 25.13 $\pm$ 0.67 & 34.01 $\pm$ 0.74 & 40.88 $\pm$ 0.81 & 46.15 $\pm$ 0.86 \\
\bottomrule
\end{tabular}
\end{table}
}

\newcommand{\gemmaninebcoding}{
\begin{table}[ht!]
\centering
\caption{Results for \gemmanineb\ on the Coding benchmark.}
\label{tab:gemma-9b-coding}
\begin{tabular}{lccccc}
\toprule
\gemmanineb & \texttt{k\_eval=1} & \texttt{k\_eval=2} & \texttt{k\_eval=4} & \texttt{k\_eval=8} & \texttt{k\_eval=16} \\
\toprule
\texttt{k\_opt=1} & \textbf{37.71} $\pm$ 0.60 & 42.03 $\pm$ 0.65 & 48.19 $\pm$ 0.69 & 55.07 $\pm$ 0.75 & 60.98 $\pm$ 0.79 \\
\texttt{k\_opt=2} & 36.84 $\pm$ 0.61 & \textbf{46.56} $\pm$ 0.67 & 52.68 $\pm$ 0.72 & 59.73 $\pm$ 0.78 & 65.86 $\pm$ 0.84 \\
\texttt{k\_opt=4} & 36.49 $\pm$ 0.63 & 44.95 $\pm$ 0.69 & \textbf{57.09} $\pm$ 0.76 & 63.64 $\pm$ 0.83 & 69.51 $\pm$ 0.88 \\
\texttt{k\_opt=8} & 35.75 $\pm$ 0.67 & 44.41 $\pm$ 0.73 & 55.08 $\pm$ 0.80 & \textbf{65.56} $\pm$ 0.88 & \textbf{71.91} $\pm$ 0.94 \\
\midrule
\cite{tangloo} & 34.36 $\pm$ 0.72 & 42.81 $\pm$ 0.78 & 52.36 $\pm$ 0.86 & 61.07 $\pm$ 0.95 & 66.41 $\pm$ 1.01 \\
\entropyreg & 35.91 $\pm$ 0.70 & 43.75 $\pm$ 0.76 & 53.28 $\pm$ 0.84 & 60.13 $\pm$ 0.93 & 65.29 $\pm$ 0.99 \\
\bottomrule
\end{tabular}
\end{table}
}

\newcommand{\llamaeightbcoding}{
\begin{table}[ht!]
\centering
\caption{Results for \llamaeightb\ on the Coding benchmark.}
\label{tab:llama-31-8b-coding}
\begin{tabular}{lccccc}
\toprule
\llamaeightb & \texttt{k\_eval=1} & \texttt{k\_eval=2} & \texttt{k\_eval=4} & \texttt{k\_eval=8} & \texttt{k\_eval=16} \\
\toprule
\texttt{k\_opt=1} & \textbf{67.38} $\pm$ 0.72 & 67.45 $\pm$ 0.76 & 69.22 $\pm$ 0.80 & 71.11 $\pm$ 0.84 & 72.84 $\pm$ 0.88 \\
\texttt{k\_opt=2} & 64.91 $\pm$ 0.73 & \textbf{69.73} $\pm$ 0.78 & 72.03 $\pm$ 0.82 & 74.08 $\pm$ 0.87 & 75.89 $\pm$ 0.91 \\
\texttt{k\_opt=4} & 64.25 $\pm$ 0.75 & 68.47 $\pm$ 0.80 & \textbf{74.67} $\pm$ 0.85 & 75.01 $\pm$ 0.90 & 77.75 $\pm$ 0.95 \\
\texttt{k\_opt=8} & 63.57 $\pm$ 0.78 & 68.39 $\pm$ 0.83 & 72.84 $\pm$ 0.88 & \textbf{76.82} $\pm$ 0.94 & \textbf{79.33} $\pm$ 0.99 \\
\midrule
\cite{tangloo} & 62.77 $\pm$ 0.82 & 66.86 $\pm$ 0.88 & 70.83 $\pm$ 0.94 & 73.95 $\pm$ 1.01 & 75.47 $\pm$ 1.06 \\
\entropyreg & 63.91 $\pm$ 0.80 & 67.85 $\pm$ 0.86 & 71.78 $\pm$ 0.92 & 72.99 $\pm$ 0.98 & 74.31 $\pm$ 1.04 \\
\bottomrule
\end{tabular}
\end{table}
}

\newcommand{\gemmaninebarc}{
\begin{table}[t!]
\centering
\caption{Results for \gemmanineb\ on the \arcagione\ benchmark.}
\label{tab:gemma-9b-arc}
\begin{tabular}{lccc}
\toprule
\gemmanineb & Cumulative Solve Rate & \texttt{pass@1} & \texttt{pass@16} \\
\toprule
\texttt{k\_opt=1} & 12.00 $\pm$ 04.33 & 02.00 $\pm$ 01.69 & 08.18 $\pm$ 04.00 \\
\texttt{k\_opt=4} & \textbf{82.33} $\pm$ 04.14 & \textbf{22.00} $\pm$ 02.00 & \textbf{38.18} $\pm$ 04.67 \\
\texttt{k\_opt=8} & \textbf{84.14} $\pm$ 04.67 & \textbf{26.67} $\pm$ 02.50 & \textbf{44.50} $\pm$ 04.33 \\
\midrule
\cite{tangloo} & 22.00 $\pm$ 04.44 & 06.00 $\pm$ 02.67 & 10.16 $\pm$ 04.57 \\
\entropyreg & 24.67 $\pm$ 04.50 & 04.00 $\pm$ 02.33 & 08.89 $\pm$ 04.89 \\
\bottomrule
\end{tabular}
\end{table}
}

\newcommand{\llamaeightbarc}{
\begin{table}[t!]
\centering
\caption{Results for \llamaeightb\ on the \arcagione\ benchmark.}
\label{tab:llama-31-8b-arc}
\begin{tabular}{lccc}
\toprule
\llamaeightb & Cumulative Solve Rate & \texttt{pass@1} & \texttt{pass@16} \\
\toprule
\texttt{k\_opt=1} & 22.00 $\pm$ 04.18 & 03.33 $\pm$ 02.00 & 08.00 $\pm$ 04.50 \\
\texttt{k\_opt=4} & \textbf{87.17} $\pm$ 04.14 & \textbf{24.33} $\pm$ 02.33 & \textbf{42.00} $\pm$ 04.16 \\
\texttt{k\_opt=8} & \textbf{88.89} $\pm$ 04.33 & \textbf{29.67} $\pm$ 02.67 & \textbf{43.13} $\pm$ 04.67 \\
\midrule
\cite{tangloo} & 36.00 $\pm$ 02.50 & 08.00 $\pm$ 04.00 & 18.00 $\pm$ 04.89 \\
\entropyreg & 28.00 $\pm$ 04.44 & 08.00 $\pm$ 02.50 & 14.67 $\pm$ 04.44 \\
\bottomrule
\end{tabular}
\end{table}
}

%% file: figures_tex/passgktoy.tex
\begin{figure}[t]
    \hfill
    \begin{subfigure}[t]{0.49\textwidth}
        \centering
        \includegraphics[width=0.99\textwidth]{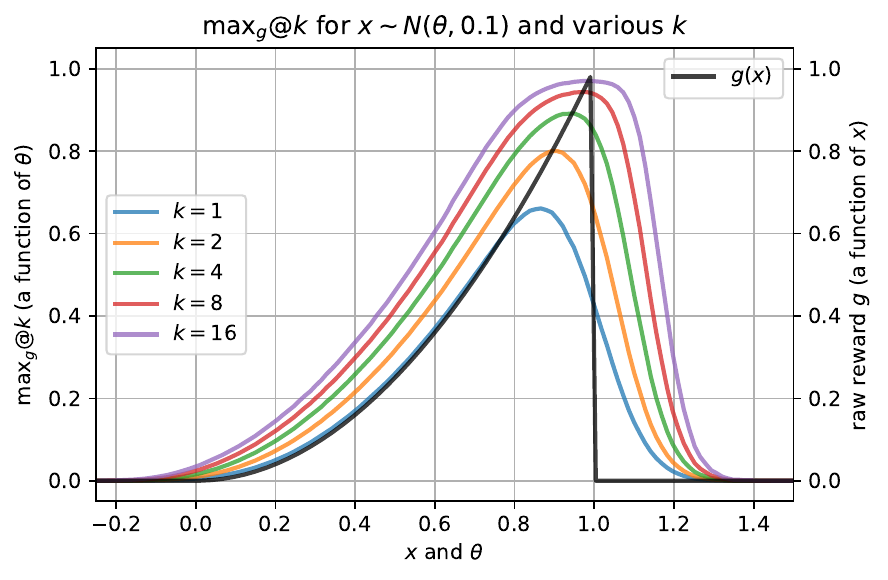}
        \caption{$g(x)$ and \maxgk.}
    \end{subfigure}
    \hfill
    \hfill
    \begin{subfigure}[t]{0.49\textwidth}
        \centering
        \includegraphics[width=0.99\textwidth]{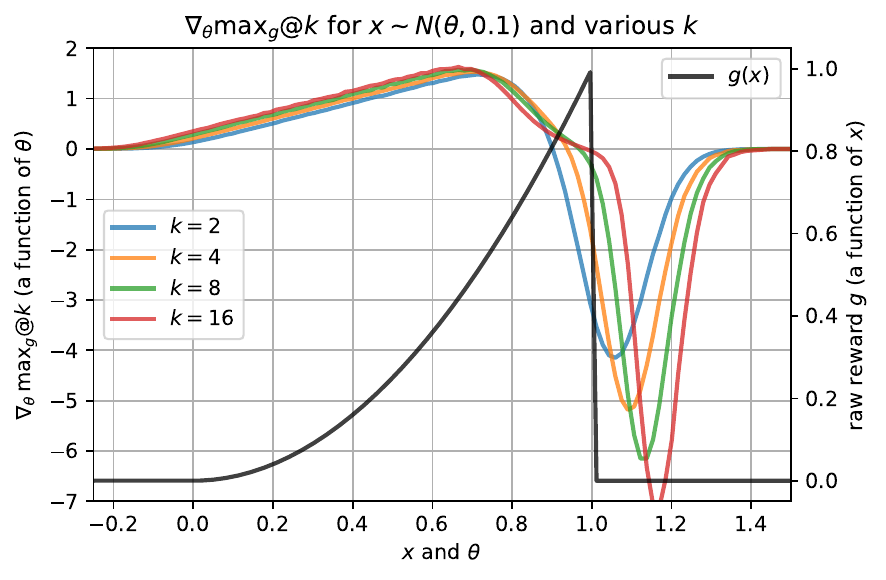}
        \caption{$g(x)$ and $\nabla_\theta \maxgk$.}
    \end{subfigure}
    \hfill
    \hfill
    \caption{The effect of $k$ on the optimal policy for a one-dimensional toy problem. The policy is normal with mean parameter $\theta$ and fixed standard deviation $0.1$. For the \maxgk\ objective (left, defined in \cref{eqn:def:passgk}) the optimal $\theta$ corresponds to the horizontal position with maximum \maxgk . For the derivative (right, the estimation of which is the focus of this paper), the optimal $\theta$ corresponds to the location of the zero crossing. For larger $k$ the optimal $\theta$ is more risk tolerant, allowing more samples to exceed one (getting zero reward) in order to increase the chance of obtaining at least one sample close to, but less than one (getting a large reward). See \cref{sec:toyexperiments} for more details.}
    \label{fig:passgktoy}
\end{figure}

%% file: figures_tex/passgkm.tex
\begin{figure}
    \centering
    \includegraphics[width=0.6\textwidth]{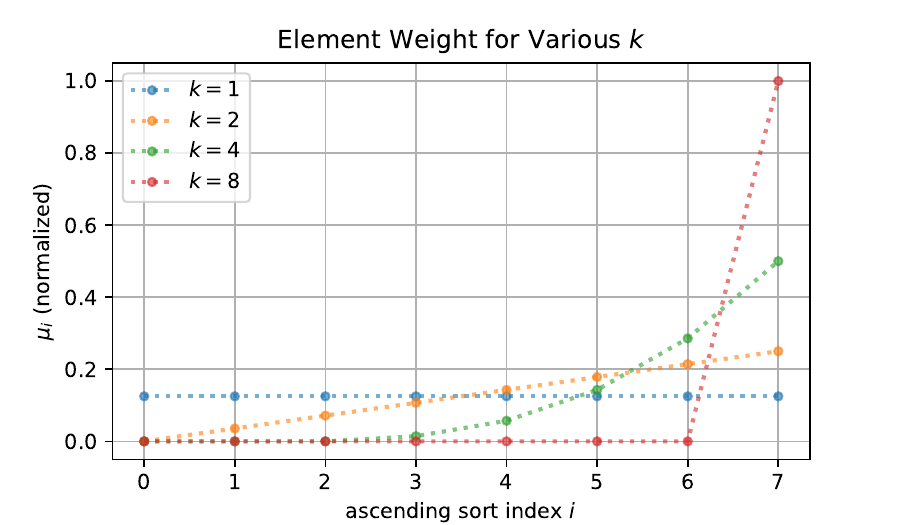}
    \caption{The effect $k$ has on the effective weight $\mu_i / \binom{n}{k}$ of \eqref{eqn:g:estsum} for a mini-batch of size $n=8$. This is the weight of the contribution of each sample assuming that the samples have been sorted in ascending order from left to right. The horizontal axis is the sort index. For $k=n=8$ only the largest sample is included; for $k=1$ all samples are weighted equally. Intermediate values interpolate these extremes in a precise manner that gives rise to unbiased gradient estimation.
    \label{fig:passgkm}
    }
\end{figure}

%% file: figures_tex/anneal.tex
\newcommand\widthannealk{0.32}
\begin{figure}[t!]
    \begin{subfigure}[t]{\widthannealk\textwidth}
        \centering
        \includegraphics[width=\textwidth]{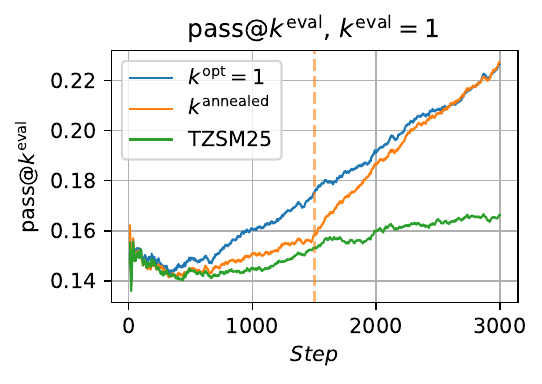}
        \caption{$\mathrm{pass}@1$ ($k^\mathrm{eval}=1$)}
        \label{fig:annealk_passone}
    \end{subfigure}
    \hfill
    \begin{subfigure}[t]{\widthannealk\textwidth}
        \centering
        \includegraphics[width=\textwidth]{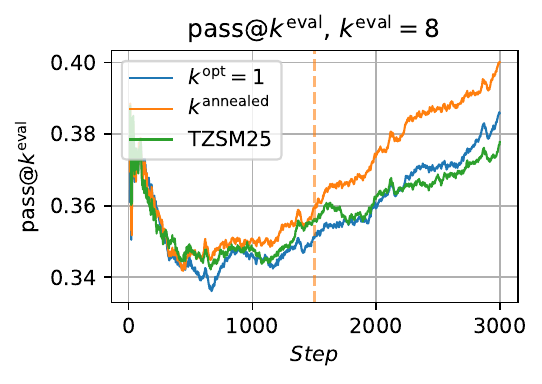}
        \caption{$\mathrm{pass}@8$ ($k^\mathrm{eval}=8$)}
    \end{subfigure}
    \begin{subfigure}[t]{\widthannealk\textwidth}
        \centering
        \includegraphics[width=\textwidth]{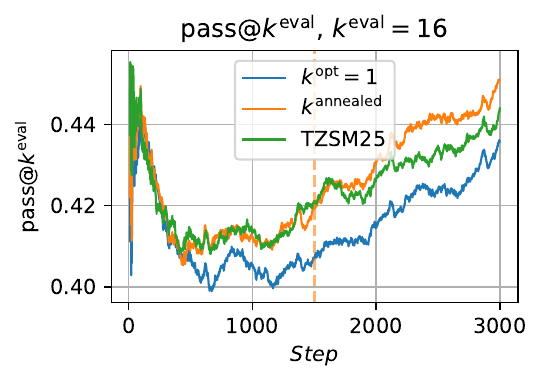}
        \caption{$\mathrm{pass}@16$ ($k^\mathrm{eval}=16$)}
    \end{subfigure}
    \hfill
    \caption{Annealing $k^\mathrm{opt}$ during PKPO training improves $\mathrm{pass}@k^\mathrm{eval}$ without sacrificing $\mathrm{pass}@1$. For \kannealed\ , we train with $k^\mathrm{opt}=8$ up to step $1500$ and $k^\mathrm{opt}=1$ thereafter.}
    \label{fig:annealk}
\end{figure}

%% file: ngeqkthm.tex
\subsection{Statement and proof that \texorpdfstring{$n\geq k$}{n >= k} samples are required to unbiasedly estimate \passk}

This result is a direct consequence of a well-known theorem concerning the unbiased estimability of parametric functions for the Bernoulli distribution.

\begin{theorem}[Kolmogorov \cite{kolmogorov}]
\label{thm:kolmog}
Let $Y_1, \dots, Y_n$ be i.i.d. Bernoulli random variables with success probability $p \in [0, 1]$. A function $\rho(p)$ is unbiasedly estimable from this sample if and only if it can be expressed as a polynomial in $p$ of degree at most $n$.
\end{theorem}

A sketch of a proof of \cref{thm:kolmog} can be found in Lehmann and Casella \cite{lehmancasella}.

\begin{corollary}
\label{thm:ngeqkforunbiasedness}
Given a sequence of $n$ i.i.d. model samples $x_1, x_2, \dots , x_n$, the \passk\ is unbiasedly estimable if and only if $n\geq k$.
\end{corollary}
\proof{
It is sufficient to consider a single a fixed and observed correctness function $f$, so that the independence of the $x_i$ implies the independence of the correctness events $[f(x_i)=1]$. Let $p=\mathbb{P}\big[[f(x_i)=1]\big]$ be the probability that any single sample is correct. The $\passk$ is defined as the complement of the probability that all $k$ samples are incorrect, which for the specific assumptions adopted in this proof is $1-(1-p)^k$. Because this expression is a polynomial in $p$ of degree $k$, the result follows immediately from \cref{thm:kolmog}.
\qed
}

%% file: variance.tex
\subsection{Characterization of the Variance}

Our proof of \cref{thm:unbiasedrho} identifies the pass@k estimator $\rho(n,c,k)$ as a $U$-statistic. To characterize its variance, we apply Hoeffding's asymptotic theory.

\begin{theorem}[Hoeffding \cite{hoeffding1948}]
\label{thm:hoeffding}
Let $X_1, \ldots, X_n$ be independent and identically distributed random variables with distribution $F$. Let $h(x_1, \ldots, x_k)$ be a symmetric kernel with $\mathbb{E}[h(X_1, \ldots, X_k)^2] < \infty$. Define the parameter $\mu = \mathbb{E}_F[h(X_1, \ldots, X_k)]$ and the $U$-statistic:
\begin{align}
U_n = \binom{n}{k}^{-1} \sum_{1 \le i_1 < \dots < i_k \le n} h(X_{i_1}, \ldots, X_{i_k}).
\end{align}
Let $h_1(x) = \mathbb{E}[h(x, X_2, \ldots, X_k)]$ be the projection of the kernel onto a single variable. Hoeffding proved that if $\zeta_1 = \text{Var}(h_1(X_1)) > 0$, then as $n \to \infty$:
\begin{align}
\sqrt{n}(U_n - \mu) \xrightarrow{d} \mathcal{N}(0, k^2 \zeta_1).
\end{align}
\end{theorem}

In the standard application of \passk\ we evaluate the estimator on a specific problem defined by a prompt and a correctness oracle. While the true pass rate $\nu$ is unknown to the observer, it is a fixed property of the model-problem pair. Consequently, the correctness outcomes of the generated samples are i.i.d. conditioned on the problem. 

The following lemma derives the variance parameter $\zeta_1$ under this conditioning. We abuse the notation by allowing the $X_i$ to denote correctness.

\begin{lemma}[Conditional Variance of the Projection]
\label{lemma:zeta_derivation}
Fix a problem instance such that the correctnesses $X_i$ are i.i.d. $\text{Bernoulli}(\nu)$. For the \passk\ kernel $h(x_1, \dots, x_k) = \max(x_1, \dots, x_k)$, the variance of the first-order projection is:
\begin{align}
    \zeta_1(\nu, k) = \nu (1-\nu)^{2k-1}.
\end{align}
\end{lemma}

\proof{
The projection $h_1(x)$ is the expected value of the kernel given the first sample is fixed to $x$, while $X_2, \dots, X_k$ remain random variates drawn from $\text{Bernoulli}(\nu)$.
\begin{align}
    h_1(x) = \mathbb{E}[ \max(x, X_2, \ldots, X_k) ].
\end{align}
We evaluate this for the two possible realizations of $x$:
\begin{enumerate}
    \item \textbf{Case $x=1$ (Success):} The maximum is 1 regardless of the remaining samples. 
    \begin{align*}
    h_1(1) = 1.
    \end{align*}
    \item \textbf{Case $x=0$ (Failure):} The maximum is 0 if and only if all remaining $k-1$ samples fail. Since the remaining samples are i.i.d. with failure probability $(1-\nu)$,
    \begin{align*}
    h_1(0) = 1 - (1-\nu)^{k-1}.
    \end{align*}
\end{enumerate}
The projection $h_1(X_1)$ is thus a binary random variable taking value $h_1(1)$ with probability $\nu$ and $h_1(0)$ with probability $1-\nu$, so that
\begin{align*}
    \zeta_1 &= \nu(1-\nu) \left( h_1(1) - h_1(0) \right)^2 \\
            &= \nu(1-\nu) \left( 1 - [1 - (1-\nu)^{k-1}] \right)^2 \\
            &= \nu(1-\nu) \left( (1-\nu)^{k-1} \right)^2 \\
            &= \nu (1-\nu)^{2k-1}.
\end{align*}
\qed
}

We can now substitute this explicit form back into Hoeffding's general result.

\begin{corollary}
\label{thm:variance}
For a fixed problem with pass rate $\nu$, as $n\rightarrow \infty$, the asymptotic variance of the estimator $\rho(n, c, k)$ is:
\begin{align}
    \text{Var}(\rho) \approx \frac{1}{n} \left[ k^2 \nu (1-\nu)^{2k-1} \right].
\end{align}
\end{corollary}

%% file: binaryproof.tex
\subsection{Proof of \texorpdfstring{\cref{thm:unbiasedrhograd}}{theorem}}
\label{sec:binaryproof}
Although \cref{thm:unbiasedrhograd} is a special case of \cref{thm:sunbiased}, we include both because the following proof uses a different approach from that of the more general statement, and is arguably the easier of the two.
\proof{
By \cref{thm:pg} the gradient $\nabla_\theta \passk$ has the unbiased estimator
\begin{align}
    \widehat \nabla 
    & \equiv
    \rho(n,c,k) \nabla_\theta \sum_{i=1}^n \log p(x_i|\theta)
    \\
    \label{eqn:substitutesuma}
    & =
    \frac{1}{\binom{n}{k}} \sum_{\substack{|\mathcal I| = k \\ \mathcal I \subseteq \{1,2,\dots,n\}}} \left(1-\prod_{i\in\mathcal I} (1-f_i)\right)
    \sum_{i=1}^n \nabla_\theta \log p(x_i|\theta)
    \\
    \label{eqn:implicitm}
    & \stackrel{\mathbb E}{\equiv}
    \frac{1}{\binom{n}{k}} \sum_{i=1}^n m_i \nabla_\theta \log p(x_i|\theta),
\end{align}
where \eqref{eqn:substitutesuma} substitutes the l.h.s. of \eqref{eqn:suma}. $m_i$ is the number of subsets $\mathcal I$ of $\{1,2,\dots,n\}$ that
\begin{enumerate}
    \item are of size $k$,
    \item contain at least one correct element, so that $\left(1-\prod_{i\in\mathcal I} (1-f_i)\right)=1$,
    \item contain $i$, so that \eqref{eqn:implicitm} holds in expectation by \cref{thm:constantpg}.
\end{enumerate}
Due to the second condition, $m_i$ therefore equals one of two values, which we denote by $m^{(1)}$ and $m^{(0)}$, depending on whether $f_i=1$ or $f_i=0$,  respectively.

If $f_i=1$ then all subsets that include $i$ also include at least one correct element ($i$ itself), so that $m^{(1)}$ is just the number of subsets of size $k$ of $\{1,2,\dots,n\}$ that include $i$, which equals the number of subsets of size $k-1$ of $\{1,2,\dots,n-1\}$:
\begin{align}
m^{(1)}=\binom{n-1}{k-1}.
\end{align}

If $f_i=0$ then we assume w.l.o.g. that $i=n$, so that $m^{(0)}$ is the number of subsets of size $k-1$ of $\{1,2,\dots,n-1\}$ with at least one correct element,
\begin{align}
    m^{(0)}
     = 
    \sum_{\substack{\mathcal J \subseteq \{1,2,\dots,n-1\} \\ |\mathcal J | = k-1}}
    \left(1-\prod_{j\in \mathcal{J} }(1-f_j)\right)
    \equiv
    \binom{n-1}{k-1} \rho(n-1, c, k-1),
\end{align}
where we again used \eqref{eqn:suma}, this time to get an expression in terms of $\rho$.
Using $m^{(0)}$ and $m^{(1)}$ we can compute $r^{(0)}$ and $r^{(1)}$ using \eqref{eqn:implicitm} as
\begin{align}
    r^{(1)} 
    & =
    \frac{m^{(1)}}{\binom{n}{k}}
     =
    \frac{\binom{n-1}{k-1}}{\binom{n}{k}}
     = \frac k n,
\end{align}
and
\begin{align}
    r^{(0)} 
    & =
    \frac{m^{(0)}}{\binom{n}{k}}
     =
    \frac{\binom{n-1}{k-1} \rho(n-1, c, k-1)}{\binom{n}{k}}
     =
    \frac k n \cdot \rho(n-1, c, k-1),
\end{align}
in line with \eqref{eqn:g:estimator}.
\qed
}

%% file: passgktimeproof.tex
\subsection{Proof of \texorpdfstring{\cref{thm:stime}}{theorem}}
\label{sec:passgktimeproof}
\proof{
The vector $\bm s = (s_1, s_2, \dots , s_n)^\top$ can be written as $\bm s = M \bm g$ where we have introduced $\bm g=(g(x_1), g(x_2), \dots , g(x_n))^\top$ as well as the matrix $M$ with
\begin{enumerate}
    \item diagonal elements $m_{ii}$ given by \eqref{eqn:g:mii},
    \item upper diagonals $m_{ij}$ for $i<j$ given by \eqref{eqn:g:mij} which is independent of $i$,
    \item lower diagonals $m_{ij}$ for $i>j$ equal to zero.
\end{enumerate}
Because of the structure of $M$, we have that
\begin{align}
    s_n 
    & = 
    \frac{1}{\binom{n}{k}} m_{nn} \, g(x_n),
\end{align}
and, for $1\leq i < n$, the right to left recursion
\begin{align}
    \label{eqn:srecursion}
    s_i = s_{i+1} + \frac{1}{\binom{n}{k}} \Big( g(x_{i})m_{ii} + g(x_{i+1})\big(m_{i,i+1}-m_{i+1,i+1}\big) \Big).
\end{align}
The ratios of $m_{ii}, m_{i,i+1}$ and $m_{i+1, i+1}$ divided by $\binom{n}{k}$ can be simplified by cancelling factors in the binomial coefficients and writing the remaining factors as a product of $k$ ratios similarly to \eqref{eqn:stablerhog}, for a total cost of $\mathcal O(nk)$; this computation can be further simplified by noting that the required ratios can be lazily computed in sequence (for example to obtain $m_{i+1,i+1}$ from $m_{ii}$) at a cost of $\mathcal O(1)$ after computing the first at a cost of $O(k)$, giving a total cost of $\mathcal O(k+n)$. The additional $\mathcal O(n\log n)$ comes from assuming the $i$ are sorted in increasing order of $g(x_i)$.
\qed
}

%% file: figures_tex/python.tex
\lstset{
  language=Python,
  commentstyle=\color[HTML]{228B22}\sffamily,
  xleftmargin=15pt
}

\begin{lstlisting}[language=Python, label=code:listing, caption={Python reward batch transformations. Functions with names that begin with an underscore are helpers, while the remaining four functions \texttt{rho}, \texttt{s}, \texttt{sloo} and \texttt{sloo\_minus\_one} implement $\rho^{(g)}$, $s_i$, $s_i^{(\mathrm{loo})}$ and $s_i^{(\mathrm{loo}-1)}$, respectively. For simplicity this implementation costs $\mathcal O(nk+n\log n)$ --- reducing this to $\mathcal O(k+n\log n)$ would require optimizing \texttt{\_deltas} and \texttt{\_m\_diagonal}.},escapechar=\#]
def _m_normed(N: int, K: int, i: int, j: int) -> float:  #\label{code:mijnormed}#
  if i == j and i >= K-1:
    return (
        K / (N-K+1) *
        np.prod(np.arange(i-K+2, i+1) / np.arange(N-K+2, N+1))
    )
  elif j > i and j >= K-1 and K >= 2:
    return (
        K / (N-K+1) * (K-1) / N *
        np.prod(np.arange(j-K+2, j) / np.arange(N-K+2, N))
    )
  return 0

def _m_diagonal(N: int, K: int) -> np.ndarray:
  return np.array([_m_normed(N, K, i, i) for i in range(N)])

def rho(g: np.ndarray, K: int) -> float:  #\label{code:rho}#
  """#\color[HTML]{228B22}See \cref{eqn:g:estsum}.#"""
  return (np.sort(g) * _m_diagonal(len(g), K)).sum()

def _delta(N: int, K: int, i: int) -> float:
  return _m_normed(N, K, i, i+1) - _m_normed(N, K, i+1, i+1)

def _deltas(N: int, K: int) -> np.ndarray:
  return np.array([_delta(N-1, K, i) for i in range(N-2)])

def _sorted_apply(func: Callable) -> Callable:
  def inner(x: np.ndarray, *args, **kwargs) -> np.ndarray:
    i_sort = np.argsort(x)
    func_x = np.zeros_like(x)
    func_x[i_sort] = func(x[i_sort], *args, **kwargs)
    return func_x
  return inner

@_sorted_apply
def s(g: np.ndarray, K: int):  #\label{code:s}#
  """#\color[HTML]{228B22}See \cref{eqn:ssum}.#"""
  N = len(g)
  c = g * _m_diagonal(N, K)
  c[:(N-1)] += g[1:] * _deltas(N+1, K)
  return np.cumsum(c[::-1])[::-1]

@_sorted_apply
def _b(g: np.ndarray, K: int) -> np.ndarray:  #\label{code:b}#
  N = len(g)
  w = (_m_diagonal(N-1, K) * np.arange(1, N)).astype(float)
  w[1:] += _deltas(N, K) * np.arange(1, N-1)
  c1 = np.array([(w * g[1:]).sum()])
  c2 = (g[:-1] - g[1:]) * w
  return np.cumsum(np.concatenate((c1, c2)))

def sloo(g: np.ndarray, K: int) -> np.ndarray:  #\label{code:sloo}#
  """#\color[HTML]{228B22}See \cref{eqn:sloo}.#"""
  return s(g, K) - _b(g, K) / (len(g) - 1)

def sloo_minus_one(g: np.ndarray, K: int) -> np.ndarray:  #\label{code:sloomo}#
  """#\color[HTML]{228B22}See \cref{eqn:sloominusone}.#"""
  return s(g, K) - _b(g, K-1) * K / (K-1) / len(g)
\end{lstlisting}

%% file: figures_tex/toyvariance.tex
\begin{figure}[h!]
    \centering
    \includegraphics[width=0.6\textwidth]{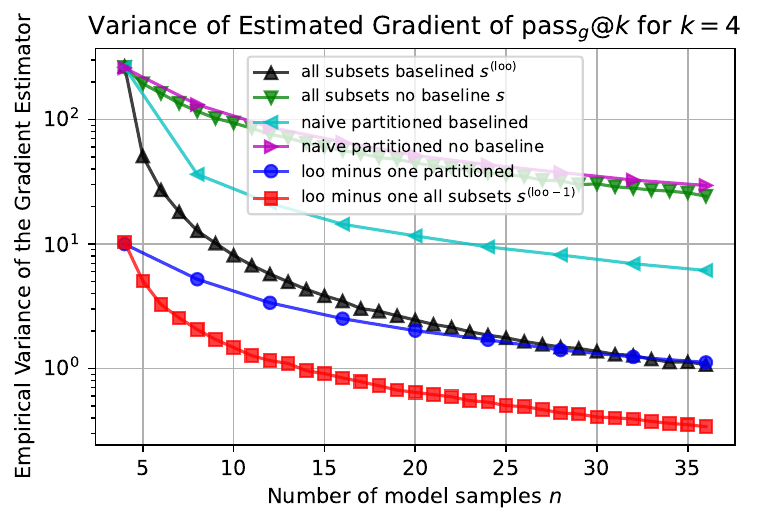}
    \caption{The variance of different estimators of the gradient of \maxgk\ with $k=4$ for the one-dimensional problem depicted in \cref{fig:passgktoy} at the location $x=1$. Each data-point is the sample variance of 10,000 independent unbiased gradient estimates (lower is better). The horizontal axis denotes the number of samples $n$ used to construct each of the 10,000 estimates. We compare the following methods: \\
    \texttt{\bf all subsets baselined: $s^{(\mathrm{loo})}$} --- our novel estimator of \cref{eqn:sloo} that analytically sums over all subsets of size $k$ of the $n$ samples with our unbiased baseline method that subtracts for each element $i$ the mean of the estimator over all subsets of size $k$ that do not include $i$. \\
    \texttt{\bf all subsets no baseline: $s$} --- our novel estimator of \cref{eqn:ssum} that analytically sums over all subsets of size $k$ of the $n$ samples but that does not include a variance-reducing baseline. \\
    \texttt{\bf naive partitioned baselined} --- a naive transformation that sets all $k$ transformed rewards in a subset of $k$ samples equal to the largest raw reward in that subset. To extend this method to $n > k$ we partition the $n$ samples (for integer multiples of $k$) into disjoint subsets of size $k$ and average the estimated gradient obtained from each. Furthermore, as a simple variance reduction method, for each such set of $k$ samples we subtract the mean of the transformed rewards from the other sets of $k$ samples (thereby averaging over $(n-k)$ samples and subtracting the result from the $k$ samples and repeating $n/k$ times in a leave-one-out fashion over the subsets of size $k$). If we were to randomly sample an increasing number of partitions of the samples and average over all of them, then intuitively the resulting estimator would approach the variance of $s^{(\mathrm{loo})}$, but this would be expensive and indeed the limiting case of considering all partitions is intractable for general $n$ and $k$. Our estimators have the key property of summing over all such partitions while nonetheless being efficient to compute. \\
    \texttt{\bf naive partitioned no baseline} --- a similar method to the previous one, but without the naive mean subtraction based variance reduction step. \\
    \texttt{\bf loo minus one partitioned} --- a method that uses the same partitioning approach as the previous two, but instead of using the naive estimate (which sets every transformed reward to simple max of the raw reward in a given set of $k$ samples) it uses the $s^{(\mathrm{loo}-1)}$ method applied separately to each disjoint set of $k$ samples, and averages that over all such subsets. In this way, this is a trivial generalization of \cite{tangloo} which extends to $n>k$ by applying the basic method to disjoint subsets and averaging the results. We do not subtract a baseline across sets as this did not improve the variance, possibly because the method within each $k$ already includes a variance reduction baseline. \\
    \texttt{\bf loo minus one all subsets: $s^{(\mathrm{loo}-1)}$} --- our novel estimator of \cref{eqn:sloominusone} that analytically sums over all subsets of size $k$ of the $n$ samples and uses all appropriate subsets of size $k-1$ to form the variance-reducing baseline that retains unbiasedness, thereby non-trivially generalizing \cite{tangloo} to all $n>k$ with strong variance reduction.
    \label{fig:toyvariance}
    }
\end{figure}

%% file: figures_tex/baseline.tex
\begin{figure}
    \centering
    \begin{subfigure}[t]{0.45\textwidth}
        \centering
        \includegraphics[width=\textwidth]{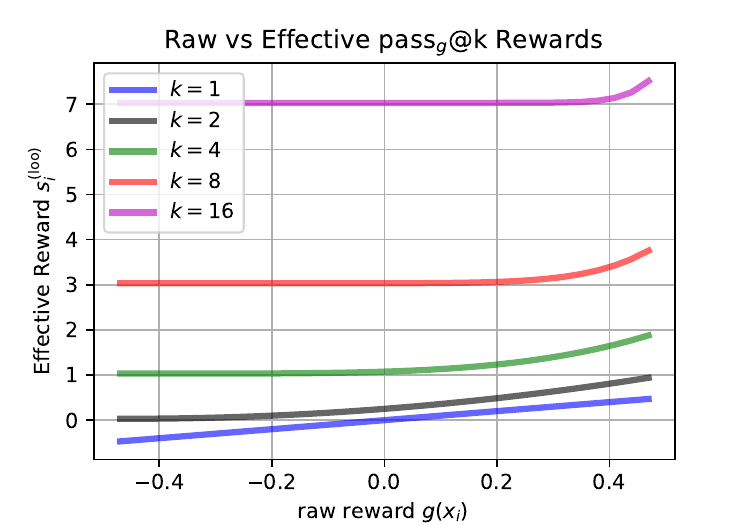}
        \caption{All subsets without baseline: $s$.}
    \end{subfigure}
    \begin{subfigure}[t]{0.45\textwidth}
        \centering
        \includegraphics[width=\textwidth]{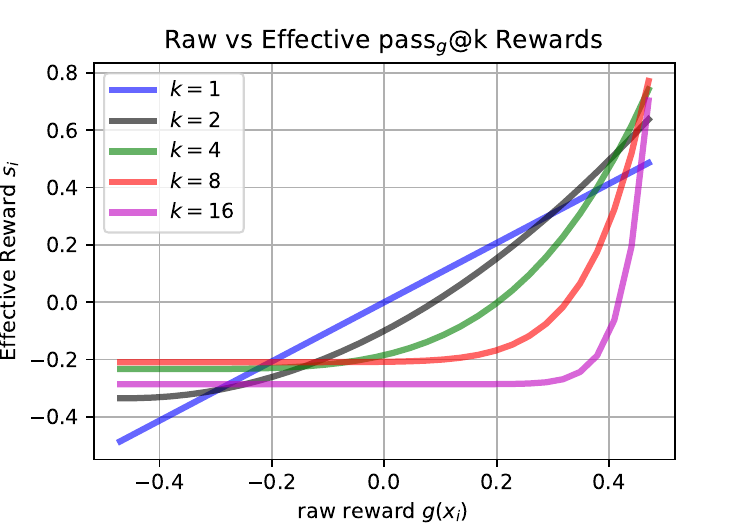}
        \caption{All subsets with LOO baseline: $s^{(\mathrm{loo})}$}
    \end{subfigure}
    \begin{subfigure}[t]{0.9\textwidth}
        \centering
        \includegraphics[width=0.5\textwidth]{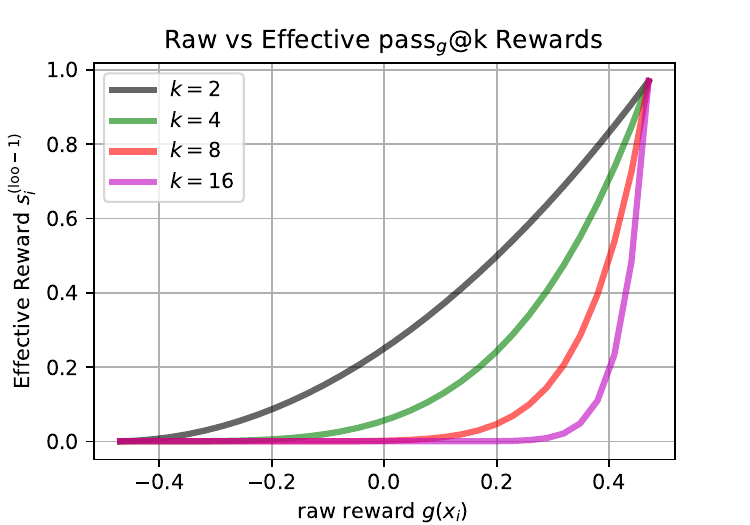}
        \caption{All $k$ sized subsets with $k-1$ sized subsets baseline: $s^{(\mathrm{loo}-1)}$}
    \end{subfigure}
    \caption{The effect of the LOO baseline on the effective rewards derived from $n=32$ raw rewards $g(x_i)$ sampled uniformly from $[-1/2, +1/2]$. The non baselined effective rewards (a) from \eqref{eqn:ssum} include a vertical offset that grows with $k$ despite being a function of raw rewards (horizontal axis) that are centered around zero. The baselined effective rewards (b) and (c) from \eqref{eqn:sloo} and \eqref{eqn:sloominusone} respectively are more centered, and give rise to reduced gradient estimator variance. To construct the figure we grouped reward values into regularly spaced bins and averaged the transformed reward for each bin to construct the curves. \textit{Note: because our transformations are from $\mathbb R^n \mapsto \mathbb R^n$ it is not possible to directly inspect a one-dimensional transformation.}}
    \label{fig:baseline}
\end{figure}

%% file: figures_tex/cumulative_solve_rate.tex
\newcommand\scale{0.99}
\begin{figure}[ht!]
    \hfill
    \hfill
    \begin{subfigure}[t]{0.49\textwidth}
        \centering
        \includegraphics[width=\scale\textwidth]{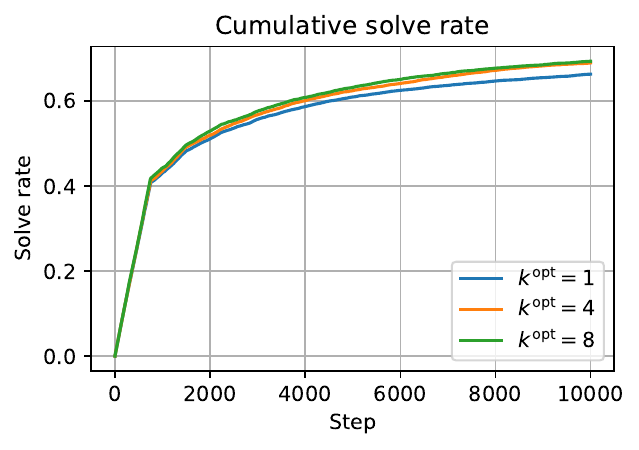}
        \caption{Cumulative solve rate ($N=12000$ problems)}
    \label{fig:cumu_solves}
    \end{subfigure}
    \hfill
    \begin{subfigure}[t]{0.49\textwidth}
        \centering
        \includegraphics[width=\scale\textwidth]{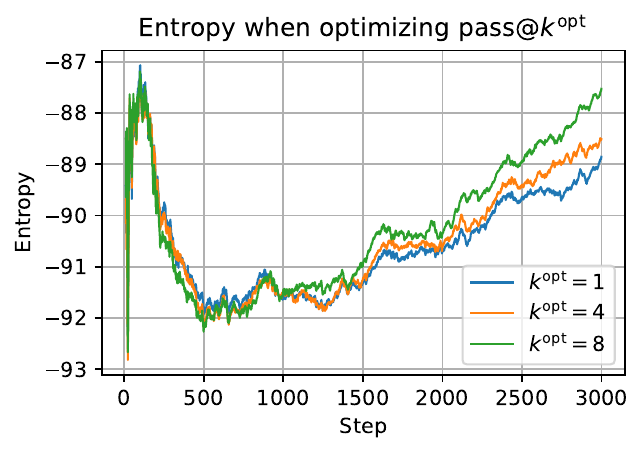}
        \caption{Effect of changing $k$ on entropy during training}
    \label{fig:entropy}
    \end{subfigure}
    \hfill
    \hfill
    \caption{(a): Increasing \kopt\ in PKPO training solves more problems during \gemma\ RL. (b): A higher \kopt\ makes the model learn to have higher entropy during RL. Thus, by optimizing for \passk\  with $k>1$ instead of \passone , the model tends to have higher entropy leading to better exploration and solving more problems. Note that the size of one epoch, which is 750 steps, is evident in (a), where we see the slope decrease at each epoch boundary.}
\end{figure}

%% file: figures_tex/rolling_pass_k.tex
\newcommand\widthpassk{0.32}
\begin{figure}[ht!]
    \begin{subfigure}[t]{\widthpassk\textwidth}
        \centering
        \includegraphics[width=\textwidth]{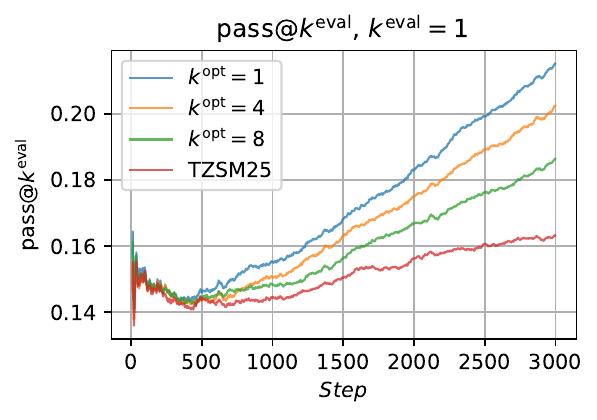}
        \caption{$\mathrm{pass}@1$ ($k^\mathrm{eval}=1$)}
    \end{subfigure}
    \hfill
    \begin{subfigure}[t]{\widthpassk\textwidth}
        \centering
        \includegraphics[width=\textwidth]{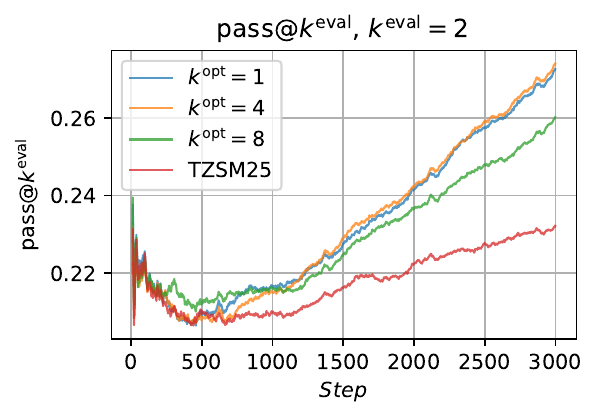}
        \caption{$\mathrm{pass}@2$ ($k^\mathrm{eval}=2$)}
    \end{subfigure}
    \hfill
    \begin{subfigure}[t]{\widthpassk\textwidth}
        \centering
        \includegraphics[width=\textwidth]{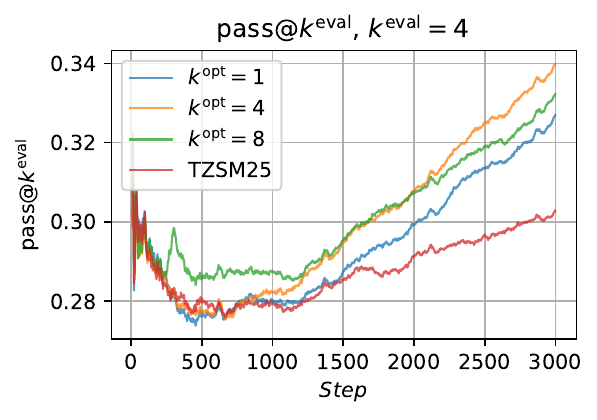}
        \caption{$\mathrm{pass}@4$ ($k^\mathrm{eval}=4$)}
    \end{subfigure}
    \hfill
    \begin{subfigure}[t]{\widthpassk\textwidth}
        \centering
        \includegraphics[width=\textwidth]{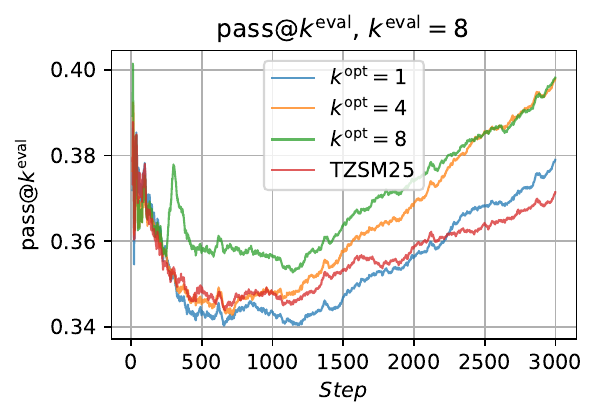}
        \caption{$\mathrm{pass}@8$ ($k^\mathrm{eval}=8$)}
    \end{subfigure}
    \hfill
    \begin{subfigure}[t]{\widthpassk\textwidth}
        \centering
        \includegraphics[width=\textwidth]{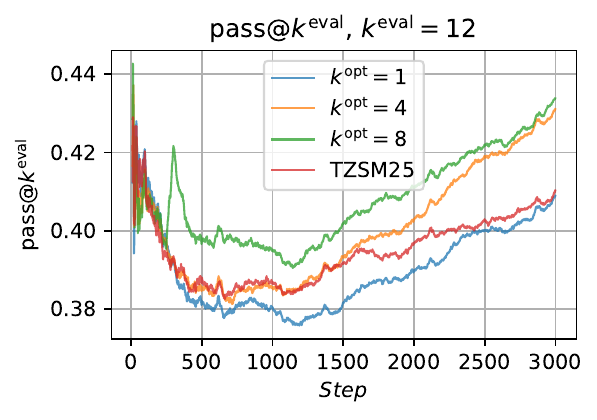}
        \caption{$\mathrm{pass}@12$ ($k^\mathrm{eval}=12$)}
    \end{subfigure}
    \hfill
    \begin{subfigure}[t]{\widthpassk\textwidth}
        \centering
        \includegraphics[width=\textwidth]{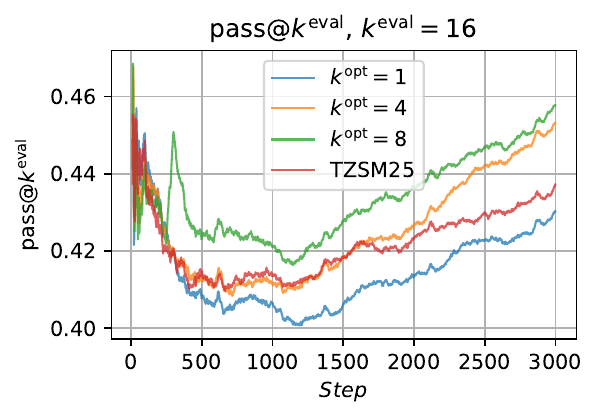}
        \caption{$\mathrm{pass}@16$ ($k^\mathrm{eval}=16$)}
    \end{subfigure}
    \caption{Effect of \kopt\ (used in our PKPO training) on the rolling \passkeval\ in \gemma\ RL. Setting $k^\mathrm{opt} = k^\mathrm{eval}$ usually achieves the best \passkeval .
    Prior work \cite{tangloo} (which is equivalent to the specific case of $k^\mathrm{opt}=n=16$ in our notation) is also shown for comparison, and suffers here presumably due to the larger estimator variance and unreliable gradient (see also \cref{fig:toyvariance}).} 
    \label{fig:rolling_passk}
\end{figure}

%% file: figures_tex/arc_agi.tex
\begin{figure}[ht!]
    \begin{subfigure}[t]{0.32\textwidth}
        \centering
        \includegraphics[width=\textwidth]{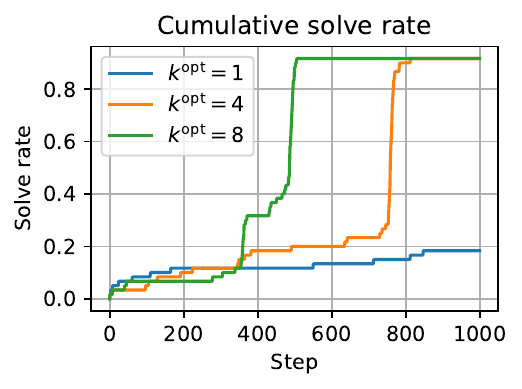}
        \caption{Cumulative solve rate}
    \end{subfigure}
    \hfill
    \begin{subfigure}[t]{0.337\textwidth}
        \centering
        \includegraphics[width=\textwidth]{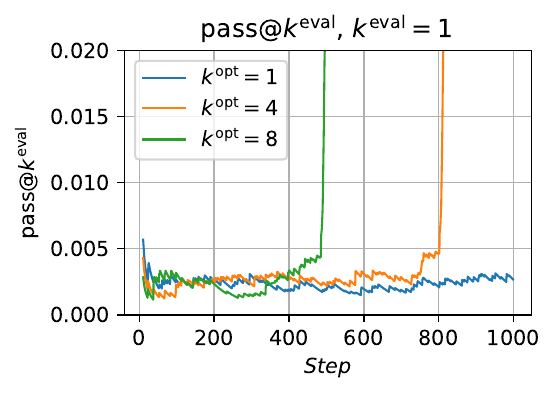}
        \caption{$\mathrm{pass}@1$ ($k^\mathrm{eval}=1$)}
    \end{subfigure}
    \hfill
    \begin{subfigure}[t]{0.33\textwidth}
        \centering
        \includegraphics[width=\textwidth]{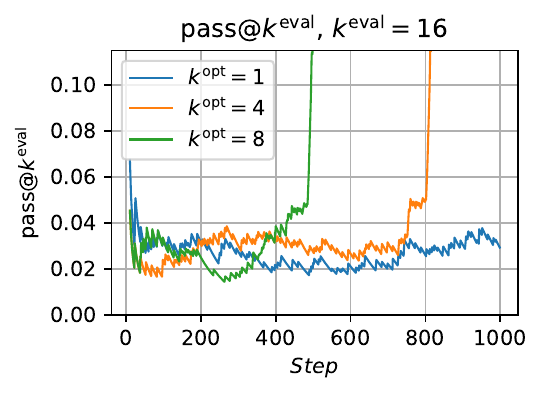}
        \caption{$\mathrm{pass}@16$ ($k^\mathrm{eval} = 16$)}
    \end{subfigure}
    \caption{Our PKPO ($k^{\mathrm{opt}}>1$) dramatically improves progress on the challenging \arcagione .}
    \label{fig:arc_agi}
\end{figure}

%% file: neurips_checklist.tex
\newpage
\section*{NeurIPS Paper Checklist}

\begin{enumerate}

\item {\bf Claims}
    \item[] Question: Do the main claims made in the abstract and introduction accurately reflect the paper's contributions and scope?
    \item[] Answer: \answerYes{} 
    \item[] Justification: the abstract and introduction are typical and represent the paper's contribution and scope.

\item {\bf Limitations}
    \item[] Question: Does the paper discuss the limitations of the work performed by the authors?
    \item[] Answer: \answerYes{} 
    \item[] Justification: The paper is largely theoretical, and the theorems include appropriate qualifiers and assumptions.

\item {\bf Theory assumptions and proofs}
    \item[] Question: For each theoretical result, does the paper provide the full set of assumptions and a complete (and correct) proof?
    \item[] Answer: \answerYes{} 
    \item[] Justification: we have made every effort to ensure that the results are precise and rigorous.

    \item {\bf Experimental result reproducibility}
    \item[] Question: Does the paper fully disclose all the information needed to reproduce the main experimental results of the paper to the extent that it affects the main claims and/or conclusions of the paper (regardless of whether the code and data are provided or not)?
    \item[] Answer: \answerYes{} 
    \item[] Justification: the main contribution is a reward transformation method that need only modify a standard RL algorithm by mapping batches of scalar rewards to their transformed values. We have provided the code for this transformation in \cref{code:listing}.

\item {\bf Open access to data and code}
    \item[] Question: Does the paper provide open access to the data and code, with sufficient instructions to faithfully reproduce the main experimental results, as described in supplemental material?
    \item[] Answer: \answerYes{} 
    \item[] Justification: the main non-trivial code that is required to transform the rewards is provided in the document as \autoref{code:listing}.

\item {\bf Experimental setting/details}
    \item[] Question: Does the paper specify all the training and test details (e.g., data splits, hyperparameters, how they were chosen, type of optimizer, etc.) necessary to understand the results?
    \item[] Answer: \answerYes{} 
    \item[] Justification: Every effort has been made to report this information to an appropriate level of detail.

\item {\bf Experiment statistical significance}
    \item[] Question: Does the paper report error bars suitably and correctly defined or other appropriate information about the statistical significance of the experiments?
    \item[] Answer: \answerYes{} 
    \item[] Justification: the paper is typical in that due to the computational cost of the experiments, we do not provide detailed condifence intervals, \textit{etc.} However, we made every effort to provide a suitable level of detail on the scope and significance of the experimental results.

\item {\bf Experiments compute resources}
    \item[] Question: For each experiment, does the paper provide sufficient information on the computer resources (type of compute workers, memory, time of execution) needed to reproduce the experiments?
    \item[] Answer: \answerYes{} 
    \item[] Justification: we have provided as much information as we are permitted to by our organization at this stage. The compute requirements are already strongly indicated by the fact that we specify that we are training with the open source 2B \gemma\ model. We can expand on this for the camera ready.
    
\item {\bf Code of ethics}
    \item[] Question: Does the research conducted in the paper conform, in every respect, with the NeurIPS Code of Ethics \url{https://neurips.cc/public/EthicsGuidelines}?
    \item[] Answer: \answerYes{} 
    \item[] Justification: The paper conforms rather comfortably, as it is mainly a theoretical paper with standard experimental results on existing publicly available data.

\item {\bf Broader impacts}
    \item[] Question: Does the paper discuss both potential positive societal impacts and negative societal impacts of the work performed?
    \item[] Answer: \answerNA{} 
    \item[] Justification: this is a methodological work that is not tied to any specific applications. There is no new and direct path from this paper to specific societal impacts.
    
\item {\bf Safeguards}
    \item[] Question: Does the paper describe safeguards that have been put in place for responsible release of data or models that have a high risk for misuse (e.g., pretrained language models, image generators, or scraped datasets)?
    \item[] Answer: \answerNA{} 
    \item[] Justification: this paper poses no such risks.

\item {\bf Licenses for existing assets}
    \item[] Question: Are the creators or original owners of assets (e.g., code, data, models), used in the paper, properly credited and are the license and terms of use explicitly mentioned and properly respected?
    \item[] Answer: \answerYes{} 
    \item[] Justification: we have made every effort to include these details.

\item {\bf New assets}
    \item[] Question: Are new assets introduced in the paper well documented and is the documentation provided alongside the assets?
    \item[] Answer: \answerNA{} 
    \item[] Justification: the paper does not introduce new assets.

\item {\bf Crowdsourcing and research with human subjects}
    \item[] Question: For crowdsourcing experiments and research with human subjects, does the paper include the full text of instructions given to participants and screenshots, if applicable, as well as details about compensation (if any)? 
    \item[] Answer: \answerNA{} 
    \item[] Justification: the paper does not use human subjects.

\item {\bf Institutional review board (IRB) approvals or equivalent for research with human subjects}
    \item[] Question: Does the paper describe potential risks incurred by study participants, whether such risks were disclosed to the subjects, and whether Institutional Review Board (IRB) approvals (or an equivalent approval/review based on the requirements of your country or institution) were obtained?
    \item[] Answer: \answerNA{} 
    \item[] Justification: see above.

\item {\bf Declaration of LLM usage}
    \item[] Question: Does the paper describe the usage of LLMs if it is an important, original, or non-standard component of the core methods in this research? Note that if the LLM is used only for writing, editing, or formatting purposes and does not impact the core methodology, scientific rigorousness, or originality of the research, declaration is not required.
    \item[] Answer: \answerNA{} 
    \item[] Justification: the LLM training in our experiments is standard and on standard publicly available tasks.

\end{enumerate}